\documentclass[10pt,journal,compsoc]{IEEEtran}

\ifCLASSOPTIONcompsoc
\usepackage[nocompress]{cite}
\else
\usepackage{cite}
\fi

\usepackage{graphicx}
\usepackage{color}
\usepackage{amsmath,amssymb}
\usepackage{multirow}
\usepackage{booktabs}
\usepackage{caption}
\usepackage{url}
\usepackage{bm}
\usepackage{color}

\ifCLASSINFOpdf
\else
\fi

\begin{document}

\title{SkeletonNet: A Topology-Preserving Solution for Learning Mesh Reconstruction of Object Surfaces from RGB Images}

\author{Jiapeng Tang,
        Xiaoguang Han,
        Mingkui Tan,
        Xin Tong,
        Kui Jia

\IEEEcompsocitemizethanks{\IEEEcompsocthanksitem J. Tang and K. Jia are with the School of Electronic and Information Engineering, South China University of Technology, the Pazhou Lab, Guangzhou, China, and also with Peng Cheng Laboratory, Shenzhen, China. E-mail: msjptang@mail.scut.edu.cn, kuijia@scut.edu.cn.
\IEEEcompsocthanksitem X. Han is with The Chinese University of Hong Kong, Shenzhen and Shenzhen Research Institute of Big Data, Shenzhen, China. E-mail: hanxiaoguang@cuhk.edu.cn.
\IEEEcompsocthanksitem M. Tan is with the School of Software Engineering, South China University of Technology, Guangzhou, China. E-mail: mingkuitan@scut.edu.cn.
\IEEEcompsocthanksitem X. Tong is with Microsoft Research Asia, Beijing, China. E-mail: xtong@microsoft.com.
\IEEEcompsocthanksitem The first two authors contributed equally to this work. K. Jia is the corresponding author.
}}

\IEEEtitleabstractindextext{
    \begin{abstract}
    This paper focuses on the challenging task of learning 3D object surface reconstructions from RGB images. Existing methods achieve varying degrees of success by using different surface representations. However, they all have their own drawbacks, and cannot properly reconstruct the surface shapes of complex topologies, arguably due to a lack of constraints on the topological structures in their learning frameworks. To this end, we propose to learn and use the topology-preserved, skeletal shape representation to assist the downstream task of object surface reconstruction from RGB images. Technically, we propose the novel \emph{SkeletonNet} design that learns a volumetric representation of a skeleton via a bridged learning of a skeletal point set, where we use parallel decoders each responsible for the learning of points on 1D skeletal curves and 2D skeletal sheets, as well as an efficient module of globally guided subvolume synthesis for a refined, high-resolution skeletal volume; we present a differentiable \emph{Point2Voxel} layer to make SkeletonNet end-to-end and trainable. With the learned skeletal volumes, we propose two models, the Skeleton-Based Graph Convolutional Neural Network (SkeGCNN) and the Skeleton-Regularized Deep Implicit Surface Network (SkeDISN), which respectively build upon and improve over the existing frameworks of explicit mesh deformation and implicit field learning for the downstream surface reconstruction task. We conduct thorough experiments that verify the efficacy of our proposed SkeletonNet. SkeGCNN and SkeDISN outperform existing methods as well, and they have their own merits when measured by different metrics. Additional results in generalized task settings further demonstrate the usefulness of our proposed methods. We have made our implementation code publicly available at {\url{ https://github.com/tangjiapeng/SkeletonNet}.}
\end{abstract}


}

\maketitle

\IEEEdisplaynontitleabstractindextext

\IEEEpeerreviewmaketitle

\begin{figure}[t]
    \vspace{-3pt}
	\centering
	\includegraphics[scale=0.33]{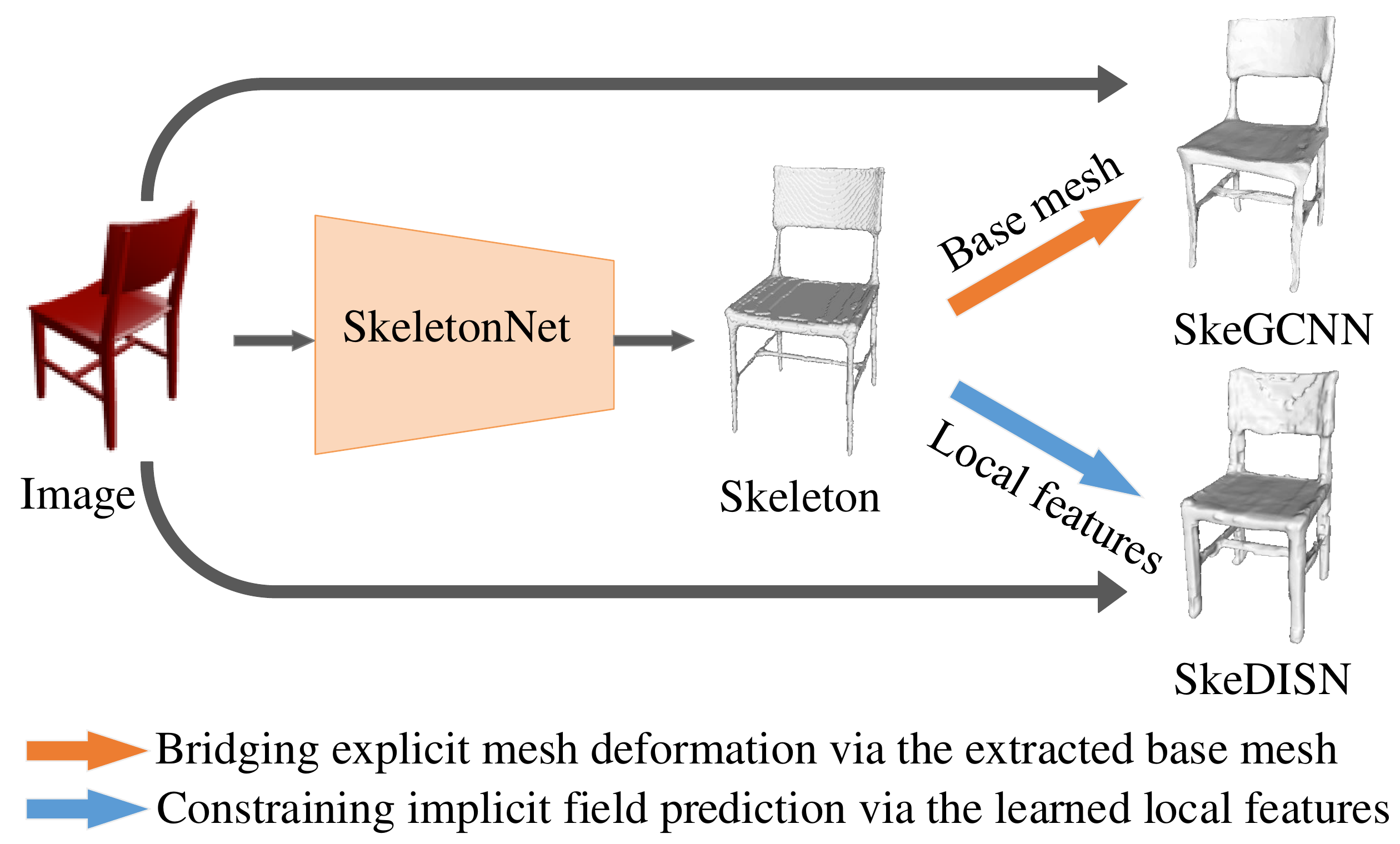}
	\caption{Given an input RGB image of an object instance, we aim to recover a surface mesh of the object and expect its topology to be correct. To tackle this challenging task, we propose \emph{SkeletonNet}, an end-to-end model that is able to efficiently generate a high-quality skeletal shape representation whose topology is the same as that of the underlying surface. We use the generated skeleton either as a bridge to \emph{explicitly} recover a surface mesh by deforming the extracted base mesh, or as a constraint to regularize the learning of an \emph{implicit} field using the learned skeleton features. We obtain state-of-the-art results using both the explicit and implicit approaches.
	}
	\vspace{-15pt}
    \label{fig:teaser}
\end{figure}

\IEEEraisesectionheading{\section{Introduction}
\label{SecIntro}}
\IEEEPARstart{R}{econstructing} surface geometries of objects or scenes from RGB images is a constituent field of computer vision research. Studies on this field of 3D vision date at least back to David Marr \cite{marr1979computational}. Classical approaches of multi-view geometry \cite{hartley2000multiple} have also been developed since then, and they serve as the foundational technologies in many applications, e.g., virtual or augmented reality, robotic navigation, and autonomous driving. While 3D surface reconstruction from multi-view images is traditionally solved by local correspondence matching \cite{hirschmuller2007stereo, campbell2008using, tola2012efficient}, followed by a global optimization \cite{merrell2007real, zach2007globally, newcombe2011kinectfusion}, it has been recently shown that, by leveraging the great modeling capacities of deep networks, the 3D surface shapes of generic objects can be learned and reconstructed from as few as a single image \cite{choy20163d,girdhar2016learning,tatarchenko2017octree,groueix2018atlasnet,kato2018neural,fan2017point,wang2018pixel2mesh,tang2019skeleton,pan2018residual,pan2019deep,michalkiewicz2019deep,mescheder2019occupancy,chen2019learning,park2019deepsdf,xu2019disn}. In spite of these preliminary successes, this inverse problem is in fact quite difficult to learn due to the arbitrary shapes of object instances and their possibly complex topologies.

Existing methods of deep learning surface reconstruction are based on different shape representations, and they range from explicit ones using point set or mesh to implicit ones using volume or continuous field. Volume-based methods \cite{choy20163d, girdhar2016learning, wu2016learning} exploit the establishment of Convolutional Neural Networks (CNNs) \cite{simonyan2015very,krizhevsky2012imagenet,szegedy2015going,he2016deep}, and simply extend CNNs as 3D versions to generate volumetric representations of surface shapes. These methods suffer from high computation and memory costs during training, which prohibit them from generating high-quality shapes represented by high-resolution volumes. This limitation is to some extent addressed by octree-based methods \cite{riegler2017octnet, riegler2017octnetfusion, tatarchenko2017octree, hane2017hierarchical, wang2017cnn, wang2018adaptive}, by designing hierarchical models based on the efficient data structures of the octree. The recent methods of deep implicit field learning \cite{chen2019learning, mescheder2019occupancy, park2019deepsdf, xu2019disn} address this issue more elegantly by learning Multi-Layer Perception (MLP)-based field functions; since the learned functions represent implicit but continuous fields of object surfaces, high-quality surface meshes can be obtained simply by querying enough points in the 3D space, followed by a final marching cubes operation \cite{lorensen1987marching}. For an explicit surface reconstruction, a deep point set regression is used in \cite{fan2017point, arsalan2017synthesizing, lin2018learning, tatarchenko2016multi} that directly regresses the coordinates of surface points. However, the discrete surface representation of point set is of lower efficiency and to improve this efficiency, point set regression is replaced in \cite{kato2018neural, groueix2018atlasnet, wang2018pixel2mesh} with the deformation of vertices in the given and initial meshes; as such, more efficient, continuous surface meshes are thus obtained. However, these explicit mesh reconstruction methods are limited by the pre-defined vertex connections that determine the topological structures of the surface to be reconstructed, and that may not be consistent with those given in the initial meshes.

Among the various challenges faced by deep learning surface reconstruction from RGB images, common to existing methods is their less efficacy to reconstruct those of complex topologies. Indeed, explicit methods have the difficulty of adjusting vertex connections, while implicit methods have no constraints on what global topological structures they would learn. To address this challenge, we connect with the \emph{skeleton}, an interior shape representation of object surface (cf. Sec. \ref{SecFramework} for the technical definition), which has the nice properties of preserving the topology of an object surface, while being of lower complexity to learn, and we expect that such a representation would be beneficial to both the implicit and explicit recoveries of surface meshes in a topology-aware manner. Technically, we propose an end-to-end, trainable system, termed \emph{SkeletonNet}, to learn skeletal shape representations from RGB images. SkeletonNet outputs skeletal volumes via the bridged learning of skeletal point sets. The skeleton is originally represented as a set of points sampled from 1D \emph{skeletal curves} or 2D \emph{skeletal sheets} \cite{tagliasacchi2012mean, wu2015deep}; we correspondingly present a parallel design of \emph{CurSkeDecoder} and \emph{SurSkeDecoder} in the SkeletonNet module of skeletal point set learning. An efficient module of globally guided subvolume synthesis is further stacked in SkeletonNet in order to obtain a high-resolution skeletal volume.
And this volumetric representation refines the skeleton by establishing an improved connectivity among skeletal points. Connecting the two modules above is a differentiable \emph{Point2Voxel} layer, which makes SkeletonNet an end-to-end model, as illustrated in Fig.~\ref{fig:skeletonNet}.

In this work, we are motivated to use this topology-preserved, skeletal shape representation to assist the downstream task of object surface recovery from RGB images. We study how to use the volumetric skeleton output of SkeletonNet either as a bridge to explicitly recover a surface mesh, or as a constraint to regularize the learning of an implicit surface field. For the former setting, we present a model of Skeleton-Based Graph CNN (SkeGCNN) that first extracts an initial mesh from the obtained skeletal volume, and then learns the weights of the GCNN to deform the vertices of the initial mesh, such that the initial mesh is inflated to fit the object surface. For the latter setting, we extend the recent state-of-the-art method of Deep Implicit Surface Network (DISN )\cite{xu2019disn}, and we present a Skeleton-Regularized DISN (SkeDISN) model that extracts multi-scale, local features from the skeletal volume to regularize the learning of a binary 3D field of object occupancy. We conduct thorough experiments that evaluate our proposed SkeletonNet for both its efficacy at producing high-quality skeletal representations and its usefulness in the downstream tasks of explicitly and implicitly learning mesh recoveries from RGB images. To enable empirical studies, we contribute a \emph{ShapeNet-Skeleton} dataset, which has prepared ground-truth skeletal point sets and skeletal volumes for object instances in ShapeNet \cite{chang2015shapenet}. Careful ablations and comparisons with various alternatives verify the importance of the components of our proposed SkeletonNet. Experiments of explicit and implicit mesh recoveries show that the use of a skeleton as a topology-preserved intermediate feature improves mesh recoveries on most object instances. It is also interesting to observe that SkeGCNN and SkeDISN outperform the respective existing methods of explicit and implicit mesh recoveries from RGB images, and they have their own merits when measured by different quantitative metrics or compared visually.

\subsection{Contributions}
Existing methods of deep learning mesh recovery from RGB images are less effective for object instances that have complex surface topologies. Our previous work \cite{tang2019skeleton} offers a preliminary attempt at using skeletons to address this limitation, where given an input RGB image, a separated, three-stage approach is used to successively predict a skeletal point set, a skeletal volume, and a final surface mesh. In the present paper, we aim for a systematic study on the usefulness of skeletal shape representations on the mesh recovery of object surfaces. To this end, we propose the end-to-end, trainable model of SkeletonNet for learning to produce skeletal shape representations. We further design models that use skeletons in the state-of-the-art frameworks of both explicit and implicit mesh recoveries from RGB images. Results of these models confirm the usefulness of skeletal representations produced by SkeletonNet. We summarize our technical contributions as follows.
\begin{itemize}
\item We propose the novel design of \emph{SkeletonNet} to learn skeletal shape representations from RGB images. SkeletonNet stacks two modules that can efficiently produce high-quality skeletal volumes via a bridged learning of skeletal point sets. We connect the two modules by a differentiable Point2Voxel layer that makes SkeletonNet end-to-end and trainable.
\item To study the usefulness of the obtained skeletal volumes, we propose two models, the Skeleton-Based Graph CNN (SkeGCNN) and the Skeleton-Regularized Deep Implicit Surface Network (SkeDISN), in the respective state-of-the-art frameworks of explicit and implicit mesh recoveries from RGB images. The former model learns to recover a surface mesh by inflating an initial mesh extracted from the skeletal volume, and the latter model regularizes the learning of a binary occupancy field using multi-scale, local features of the skeletal volume.
\item We conduct thorough experiments that confirm both the efficacy of SkeletonNet at producing high-quality skeletal representations, as well as its usefulness for downstream surface recovery tasks. By comparing SkeGCNN and SkeDISN with existing methods, the respective advantages of explicit and implicit methods under different measures manifest themselves as well. Experiments under generalized settings, including deep learning mesh recovery from multi-view images, further demonstrate the usefulness of our proposed methods.
\end{itemize}

\section{Related Works}
\label{SecRelatedWork}

In this section, we review existing methods of deep learning surface reconstruction from RGB images. We focus on the mesh representation of object surfaces, while other representations, e.g., point sets or volumes, are briefly discussed as well. We organize this review into two lines of methods that can either explicitly learn a deformation of mesh vertices or learn an implicit surface representation from which the surface mesh can be extracted. We finally discuss existing research where the interior shape representation of the skeleton is used for various downstream tasks.

\vspace{0.1cm}
\noindent\textbf{Explicit Surface Learning}
Point set is a discrete and simple, and arguably the most popular representation of object surface. Deep learning surface reconstruction starts from recovery of the surface point set. For example, Fan et al. \cite{fan2017point} propose learning a deep regression model for generating coordinates of points residing on the surface of an object observed in a single-view image, where the Chamfer distance (CD) or the Earth Mover's distance (EMD) between point sets are used to build the training objective. To improve the fidelity of the generated point sets, adversarial training is adopted by Jiang et al. \cite{jiang2018gal} and Achlioptas et al. \cite{achlioptas2018learning} to constrain the distribution of the learned latent space. Our proposed SkeletonNet is also concerned with an intermediate prediction of skeletal point set, for which we build upon existing works \cite{groueix2018atlasnet} but introduce novel designs as well, in order to better generate skeletal points whose neighboring relations form 1D skeletal curves and 2D skeletal sheets.

\noindent Meshes are efficient and continuous surface representations that connect isolated points, i.e. mesh vertices, with mesh edges and faces. Learning to reconstruct a surface mesh directly is challenged by the learning of connecting relations among mesh vertices. To alleviate this issue, existing methods ~\cite{kato2018neural,wang2018pixel2mesh,pan2019deep} convert this challenge to learning the vertex deformation of initial meshes. For example, AtlasNet~\cite{kato2018neural} and DGP~\cite{Williams_2019_CVPR} propose to learn deep networks to deform multiple, initial meshes and expect the deformed ones to cover the surface altogether; as such, they fail to produce a closed surface mesh. Pix2Mesh \cite{wang2018pixel2mesh} is able to obtain a closed mesh by learning to deform vertices of a single, initial mesh; however, given the fixed topology of the initial mesh, it cannot learn to generate surface meshes whose topologies are unknown. Although some methods \cite{jack2018learning, pontes2018image2mesh, wang20193dn} search for initial mesh structures from a set of reference models, they are still limited when it comes to reconstruct the surface meshes of complex topologies. More recently, Pan et al. \cite{pan2019deep} present the topology modification network (TMNet) to learn and update mesh connectivity when deforming the initial mesh, where topology modification is achieved by a mechanism of face pruning based on estimated face-to-surface distances. This is further improved by Nie et al. \cite{nie2020total3dunderstanding} with an adaptive thresholding strategy. However, learning the relations of vertex connections dynamically during the deformation process is a rather challenging task, causing these approaches to be less effective for the recovery of thin structures (cf. the examples in Fig. \ref{fig:comp_recon}).

\noindent The present work acknowledges these challenges faced by existing methods. Instead of addressing these challenges directly, we propose to decouple these learning difficulties as challenges concerned with the recovery of topological structures and challenges concerned with the recovery of surface geometries. Technically, we propose SkeletonNet and expect that the produced skeletal representations can serve as topology-preserved, initial meshes to be subsequently inflated as surface meshes.
A similar strategy is also adopted in Point2Mesh \cite{hanocka2020point2mesh}, which adjusts the vertices of initial meshes to approximate the observed point clouds. However, different from directly extracting initial meshes from the volumetric representation of given point sets, it is non-trivial to obtain initial meshes with adaptive topologies from input RGB images. SkeletonNet is thus proposed to address this challenge.

\vspace{0.1cm}
\noindent\textbf{Implicit Surface Learning}
An object surface can also be represented implicitly as a discrete volume or a continuous field, from which a surface mesh can be extracted, e.g., via marching cubes \cite{lorensen1987marching}. Learning to reconstruct volumetric shapes gains earlier popularity due to the regularity of the data structures involved \cite{choy20163d, girdhar2016learning, wu2016learning, kar2017learning}. These methods simply extend the CNNs for 2D image modeling to their 3D counterparts. For example, the work of ~\cite{choy20163d} combines 3D convolutions and long short-term memory (LSTM) units to achieve volumetric grid reconstructions from single- or multi-view RGB images. A 3D auto-encoder is trained in \cite{girdhar2016learning} whose decoder is used to build the mapping from a 2D image to a 3D occupancy grid. Wu et al. \cite{wu2016learning} adopt generative adversarial networks \cite{goodfellow2014generative} to learn a better volumetric shape reconstruction. Due to the high computation cost of 3D convolution, these methods are only able to produce low-resolution volumes. The hierarchical structure of an octree would alleviate this issue to some extent \cite{riegler2017octnet, riegler2017octnetfusion, tatarchenko2017octree, hane2017hierarchical, wang2017cnn, wang2018adaptive}; however, it seems that such an octree-based data structure makes the training of compatibly designed deep models less effective, resulting in shape reconstructions that often have artifacts.

\noindent It has been recently discovered that various representations of implicit surface fields are quite effective for object surface modeling and reconstruction, with the signed distance function (SDF)~\cite{park2019deepsdf, chen2020bsp, lei2020analytic, atzmon2020sal,gropp2020implicit} and the object occupancy field~\cite{mescheder2019occupancy, chen2019learning, chibane2020implicit, peng2020convolutional} as prominent examples. For example, the works of~\cite{chen2019learning, mescheder2019occupancy, michalkiewicz2019deep, saito2019pifu} learn a 3D field of object occupancy, where an object's surface is considered as a continuous decision boundary that separates the 3D space as either inside or outside the object. The methods of~\cite{park2019deepsdf, xu2019disn} learn a function of SDF that specifies not only whether 3D points are inside or outside the object but also how far they are from the object surface.
The pioneer works~\cite{mescheder2019occupancy, chen2019learning, park2019deepsdf} choose to learn a global field from a single latent code. To preserve local shape patterns, DISN~\cite{xu2019disn}, IF-Net~\cite{chibane2020implicit}, and Conv-ONet~\cite{peng2020convolutional} learn local implicit fields based on convolutional features extracted from inputs. Most of the works rely on signed objectives during training, while SAL~\cite{atzmon2020sal} and IGR~\cite{gropp2020implicit} utilize sign-agnostic learning techniques to achieve surface reconstruction directly from un-oriented point clouds, with no use of oriented normals. Another work of DefTet~\cite{gao2020learning} chooses to deform vertices of tetrahedral meshes with predicted occupancies, which can bypass the post-processing of marching cubes for triangular mesh extraction. In addition to geometry inference, some works extend the implicit representation for appearance modeling, such as modeling textures~\cite{oechsle2019texture} or light conditions~\cite{oechsle2020learning, yariv2020multiview}.
These methods typically use MLPs as the implicit models, and learn latent shape encodings of the input RGB images via CNNs. Owning to the continuity of implicit fields, surface reconstructions from these methods are often smooth and can ideally achieve infinite resolution.  However, due to the lack of explicit constraints on what global topological structure an implicit field can learn, these methods are not guaranteed to correctly recover the global topology of an object from an input image. Our proposed use of the topology-preserved skeleton as a structural regularization is indeed intended to address this limitation.

\vspace{0.1cm}
\noindent\textbf{The Skeleton as an Interior Shape Representation}
The medial axis transform (MAT)~\cite{blum1967transformation} is an intrinsic shape representation that includes the medial axis together with the associated radius function of the maximally inscribed spheres, which can be used to reconstruct the original shape. The MAT has direct access to both a shape's boundary and its interior, enabling various downstream applications regarding shape abstraction, classification and segmentation. Here, we refer the reader to ~\cite{siddiqi2008medial}, which presents a systematic discussion on the applications of MAT. To drive these applications, many methods ~\cite{attali1996modeling, chazal2005lambda, miklos2010discrete} are designed to compute the MAT for downstream applications such as shape analysis, modeling, and editing. For example, Q-MAT~\cite{li2015q}  and Q-MAT+~\cite{pan2019q} define novel quadratic errors to conduct MAT mesh simplification so that they can compute geometrically accurate, structurally simple, and compact meshes, which are employed either as supervision to constrain the MAT learning from point clouds~\cite{yang2020p2mat} or as input to extract structural features for 3D object recognition~\cite{hu2019mat}. But in reality, clean 3D mesh models from scanners are not always approachable, which leads to great difficulties in computing accurate MAT meshes.
The skeletal shape representation, mostly as a compact point cloud, is considered as an approximation for medial axis transformation~\cite{blum1967transformation}.
 Some existing works \cite{cornea2005curve, sharf2007fly, tagliasacchi2009curve, bucksch2010skeltre, tagliasacchi2012mean, huang2013l1} address meso-skeleton extraction from point sets; a meso-skeleton~\cite{tagliasacchi2012mean} consists of a mixture of curves and surface sheets adapted to the local 3D geometry, and it is appropriate for representing the topological structures of general 3D shapes. Wu et al.~\cite{wu2015deep} apply meso-skeleton to surface reconstruction from imperfect point clouds; they extract a meso-skeleton to capture contextual geometric information, which guides the completion of missing regions on the surface. Recently, P2PNet~\cite{yin2018p2p} and Nie et al.~\cite{nie2020skeleton} introduce learning-based approaches to transform surfaces into meso-skeletons. Point2Skeleton~\cite{lin2020point2skeleton} learns skeletal points in an unsupervised manner via a convex combination of the given point cloud. PIE-Net~\cite{wang2020pie} learns to extract parametric curves from a point cloud. Different from these methods that aim at skeleton learning from point cloud data, we focus on skeleton inference from RGB images, which is even more challenging. Our motivation is to use the topology-preserved, skeletal shape representation to assist the down-stream task of object surface recovery from RGB images. To the best of our knowledge, we are the first to explore skeleton inference from RGB images. 



\section{Overview}
\label{SecFramework}

\begin{figure*}[t]
	\begin{center}
		\includegraphics[scale=0.28]{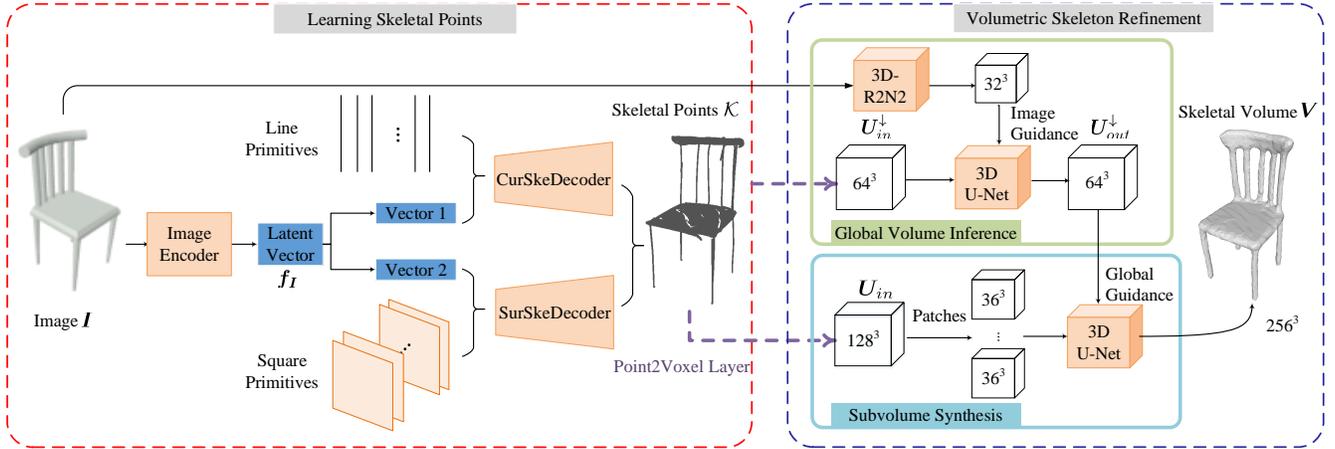}
		\caption{The pipeline of our proposed \emph{SkeletonNet}. The module in the red dotted box (left) stacks the parallel \emph{CurSkeDecoder} and \emph{SurSkeDecoder} on top of a CNN based image encoder; given an input image $\bm{I}$, the module is trained to regress a skeletal point set $\mathcal{K}$. The module in the blue dotted box (right) consists of two parallel streams of 3D CNNs; the top one is for a global, low-resolution volume synthesis, which guides the synthesis of a high-resolution skeletal volume $\bm{V}$ via a sliding subvolume fashion in the bottom stream. Connecting the two modules is a differentiable \emph{Point2Voxel} layer that converts $\mathcal{K}$ as the initial volume and makes the proposed SkeletonNet an end-to-end trainable network. }
		\label{fig:skeletonNet}
	\end{center}
\end{figure*}

Given an RGB image $\bm{I}$ of an object as input, our goal is to learn a deep model from training data, such that the object surface can be recovered as a triangle mesh $\mathcal{M}$ = \{$\mathcal{T}$, $\mathcal{E}$, $\mathcal{F}$\}, where $\mathcal{T}$ denotes the set of mesh vertices, $\mathcal{E}$ for the edges connecting adjacent vertices, and $\mathcal{F}$ for the faces defined by three adjacent vertices and their connected edges. The object surface to be recovered could have high algebraic and/or topological complexities, which pose great challenges to existing methods \cite{choy20163d, tatarchenko2017octree, fan2017point, kato2018neural, wang2018pixel2mesh, groueix2018atlasnet, mescheder2019occupancy, chen2019learning}.

 To address the challenges, our key idea in this work is to leverage the \emph{meso-skeleton} \footnote{The skeletal shape representation is a kind of approximation for the medial axis transform (MAT) \cite{blum1967transformation}. While the MAT of a 2D shape is a 1D skeleton, for a 3D model, the MAT is generally composed of 2D surface sheets. The skeleton composed of skeletal curves and skeletal sheets (i.e., medial axes) is generally called meso-skeleton ~\cite{tagliasacchi2012mean}, shortened as skeleton. } of object surface either as a bridge or as a constraint, such that a better surface recovery could be subsequently achieved. Similar to a point cloud approximation of an object surface, a skeleton is usually represented as a point set (cf. examples in Fig. \ref{fig:allcats_ske}). Given a surface shape, skeletal points can be obtained by sinking the sampled surface points along the reverse directions of their normals. These points are considered as sampled ones from \emph{skeletal curves} and \emph{skeletal sheets}, which together define the meso-skeleton of object surface that has the nice properties of topology preservation and lower learning complexity. Our motivation to use skeletons to assist surface recovery is in fact based on these properties.

\vspace{0.1cm}
\noindent\textbf{\emph{Learning Skeletons from RGB Images}}
Our system of surface recovery starts from learning a skeleton that lies in the object canonical space from the input $\bm{I}$. While the skeleton is originally represented as a point set $\mathcal{K} = \{ \bm{p}_i \in \mathbb{R}^3 \}_{i=1}^{n_{\mathcal{K}} }$, we propose a further learning to refine $\mathcal{K}$ as a skeletal volume $\bm{V} \in \mathbb{R}^{r\times r\times r}$,  as shown in Fig. \ref{fig:skeletonNet}. Learning $\bm{V}$ via 3D CNNs is beneficial to establishing the connectivity among skeletal points, from which a subsequent meshing can also be achieved directly via marching cubes~\cite{lorensen1987marching}. To learn $\mathcal{K}$, we present a novel, parallel design of \emph{CurSkeDecoder} and \emph{SurSkeDecoder} stacked on top of a shared CNN based image encoder; the CurSkeDecoder and SurSkeDecoder are designed to be respectively responsible for regression of points on skeletal curves and skeletal sheets. Given $\mathcal{K}$, we present an efficient scheme of globally guided subvolume synthesis to learn $\bm{V}$, which is essential to produce a high-resolution skeletal volume (e.g., $|\bm{V}| = 256^3$). Connecting the above two modules is a differentiable \emph{Point2Voxel} layer, which makes an end-to-end model of our proposed SkeletonNet, as illustrated in Fig. ~\ref{fig:skeletonNet}. We present specifics of SkeletonNet in Sec. ~\ref{SecSkeleonNet}.

\vspace{0.1cm}
\noindent\textbf{\emph{Skeleton-guided Surface Learning}}
As an interior representation of object surface, the skeleton $\bm{V}$ (and $\mathcal{K}$) preserves the surface topology, and is potentially helpful for learning to recover $\mathcal{M}$. In this work, we study how to learn $\mathcal{M}$ either \emph{explicitly} from the skeletal volume $\bm{V}$, or \emph{implicitly} by using features of $\bm{V}$ to regularize the learning of a surface field. For the intended explicit surface learning, we present a design of graph convolutional network that produces $\mathcal{M}$ by learning vertex deformation of an initial mesh $\mathcal{M}_{\bm{V}}$ extracted from $\bm{V}$; Fig. \ref{fig:mesh_refine_alg} gives the illustration. For learning of an implicit surface field, we improve upon recent methods \cite{xu2019disn} by augmenting their image-level features with multi-scale features from the skeletal volume $\bm{V}$, such that a probability field of occupancy \cite{mescheder2019occupancy, chen2019learning} can be better learned, from which $\mathcal{M}$ can be recovered via marching cubes \cite{lorensen1987marching}; Fig. \ref{fig:imp_mesh_alg} gives the illustration. We conduct thorough experiments that confirm the efficacy of our proposed SkeletonNet for learning surface recovery from RGB images.

\section{The Proposed SkeletonNet}
\label{SecSkeleonNet}

We present in this section specifics of our proposed SkeletonNet. Given an input $\bm{I}$,  SkeletonNet is trained in an end-to-end fashion to produce a high-resolution skeletal volume $\bm{V} \in \mathbb{R}^{r\times r\times r}$ via an intermediate representation of skeletal point set $\mathcal{K} = \{ \bm{p}_i \in \mathbb{R}^3 \}_{i=1}^{n_{\mathcal{K}} }$. While both $\mathcal{K}$ and $\bm{V}$ are skeleton representations of the underlying surface for the object contained in $\bm{I}$, they have different properties: the point set representation is original and compact, and the volumetric representation is more regular and connected. Our way of learning $\bm{V}$ via the intermediate $\mathcal{K}$ enjoys both of their advantages by establishing the connectivity among skeletal points in $\mathcal{K}$ to form skeletal curves and skeletal sheets in $\bm{V}$. We empirically find that the alternative ways of either learning $\mathcal{K}$ alone or learning $\bm{V}$ directly from $\bm{I}$ are less effective to have high-quality skeletons.

\vspace{0.1cm}
\noindent\textbf{\emph{Image Encoder}}
The SkeletonNet starts with a CNN based image encoder, which produces a latent vector $\bm{f}_{\bm{I}} \in \mathbb{R}^{m}$ encoding surface shape of the object in $\bm{I}$. $\bm{f}_{\bm{I}}$ will be used as input of the module for learning $\mathcal{K}$, as described shortly.

\subsection{Learning of Skeletal Points}
\label{SecSkeletonLearning}

 Learning $\mathcal{K}$ from the latent vector $\bm{f}_{\bm{I}}$ is a standard point set regression problem. For example, one may choose the technique in PointSetGen \cite{fan2017point} to directly regress the coordinates of skeletal points in $\mathcal{K}$. However, we note that skeletal points are those sampled from their underlying skeletal curves and skeletal sheets, and as such, distribution of these points satisfies geometric constraints that prevent them from free positioning in the 3D space. To implement such constraints, we introduce a novel, parallel design of \emph{CurSkeDecoder} and \emph{SurSkeDecoder}, which are respectively responsible for regression of skeletal points on 1D curves, denoted collectively as $\mathcal{K}_{Cur}$, and those on 2D sheets, denoted as $\mathcal{K}_{Sur}$.

Specifically, we define a set of 2D primitives of unit square $[0, 1]^2$. Except for $\bm{f}_{\bm{I}}$, SurSkeDecoder takes as input regularly sampled points from each of the unit squares, and learns to deform the sampled points such that they form points on skeletal sheets in the 3D space; in other words, SurSkeDecoder learns a function $\psi: [0, 1]^2 \rightarrow \mathbb{R}^3$. Similarly,we define a set of 1D primitives of line segment $[0, 1]$. CurSkeDecoder learns a function $\phi \in [0, 1] \times \mathbb{R}^3$ by sampling regular points from the segments, and then learning to deform the sampled points such that they form points on skeletal curves in the 3D space. This way of using multiple 1D or 2D primitives respectively as inputs of CurSkeDecoder or SurSkeDecoder is similar to the strategy in AtlasNet ~\cite{groueix2018atlasnet}. In this work, both CurSkeDecoder $\phi(\cdot)$ and SurSkeDecoder $\psi(\cdot)$ are implemented as multilayer perceptrons (MLPs), whose layer specifics are given in Sec. \ref{SecExp}. Note that neighboring relations are fixed for connected points sampled from each 1D or 2D primitive. To maintain the spatial relations among neighboring points during training, we propose the following regularized loss function to learn CurSkeDecoder $\phi(\cdot)$.
\begin{align}\label{EqnCurSkeDecoderLoss}
    L_{\phi} = D(\mathcal{K}_{Cur}, \mathcal{K}_{Cur}^{*}) + \alpha R_{laplacian}(\mathcal{K}_{Cur}) ,
\end{align}
with
\begin{align}
    D(\mathcal{K}_{Cur}, \mathcal{K}_{Cur}^{*})  = \sum_{\bm{p} \in \mathcal{K}_{Cur}} \min_{\bm{p}^{*} \in \mathcal{K}_{Cur}^{*}} \| \bm{p} - \bm{p}^{*} \|_2^2 \ \ + \nonumber \\ \quad\quad \sum_{\bm{p}^{*} \in \mathcal{K}_{Cur}^{*}} \min_{\bm{p} \in \mathcal{K}_{Cur}} \| \bm{p} - \bm{p}^{*} \|_2^2 , \label{EqnCurSkeDecoderLossCD} \\
    R_{laplacian}(\mathcal{K}_{Cur}) = \sum_{\bm{p} \in \mathcal{K}_{Cur}} \left\| \bm{p} - \frac{1}{|\mathcal{N}(\bm{p})|} \sum_{\bm{p}' \in \mathcal{N}(\bm{p})} \bm{p}' \right\|_2^2  , \label{EqnCurSkeDecoderLossLaplacian}
\end{align}
where $\alpha$ is a penalty parameter, and $\mathcal{K}_{Cur}^{*}$ denotes the ground-truth point set consisting of those on skeletal curves. We use Chamfer distance to define the loss term (\ref{EqnCurSkeDecoderLossCD}), and laplacian smoothness \cite{field1988laplacian} to define the regularizer (\ref{EqnCurSkeDecoderLossLaplacian}), where $\mathcal{N}(\bm{p})$ contains the neighboring points of any $\bm{p} \in \mathcal{K}_{Cur}$ whose pre-images are connected to that of $\bm{p}$ on a 1D unit line segment. We similarly define the following loss to learn SurSkeDecoder $\psi(\cdot)$
\begin{align}\label{EqnSurSkeDecoderLoss}
    L_{\psi} = D(\mathcal{K}_{Sur}, \mathcal{K}_{Sur}^{*}) + \alpha R_{laplacian}(\mathcal{K}_{Sur}) ,
\end{align}
where $D(\mathcal{K}_{Sur}, \mathcal{K}_{Sur}^{*})$ and $R_{laplacian}(\mathcal{K}_{Sur})$ are similarly defined as (\ref{EqnCurSkeDecoderLossCD}) and (\ref{EqnCurSkeDecoderLossLaplacian}). We note that $\mathcal{K}^{*} = \mathcal{K}_{Cur}^{*} \bigcup \mathcal{K}_{Sur}^{*}$ and $\mathcal{K}_{Cur}^{*} \bigcap \mathcal{K}_{Sur}^{*} = \emptyset$. To prepare $\mathcal{K}_{Cur}^{*}$ and $\mathcal{K}_{Sur}^{*}$, we present the ShapeNet-Skeleton dataset that contains the annotations of $\mathcal{K}_{Cur}^{*}$ and $\mathcal{K}_{Sur}^{*}$ for each object instance in ShapeNet \cite{chang2015shapenet}.

\vspace{0.1cm}
\noindent \textbf{\emph{The ShapeNet-Skeleton Dataset}}
To facilitate training of CurSkeDecoder and SurSkeDecoder, we prepare training data of skeletal points for object instances in ShapeNet \cite{chang2015shapenet} as follows: 1) for each CAD model in ShapeNet, we convert it into a point cloud; 2) we extract the point set $\mathcal{K}^{*}$ of meso-skeleton using the method \cite{wu2015deep}; 3) to have $\mathcal{K}_{Cur}^{*}$ and $\mathcal{K}_{Sur}^{*}$, we classify each $\bm{p}^{*} \in \mathcal{K}^{*}$ as either belonging to skeletal curves or belonging to skeletal sheets based on principal component analysis of its neighboring points. Specifically, a point that has only one dominating eigenvalue is classified as that of skeletal curves, and as that of skeletal sheets otherwise. We have made this dataset publicly available at {\url{https://github.com/tangjiapeng/SkeletonNet}}. Given the ground truth $\mathcal{K}^{*}$ of an object instance, we also prepare its volumetric version $\bm{V}^{*}$ by first quantizing the skeletal points in $\mathcal{K}^{*}$ as voxels and then conducting the operation of interior filling to have a solid volume. For better preserving the long and thin parts, we finally use morphological dilated skeletal volumes as ground truths.

\subsection{Volumetric Skeleton Refinement}
\label{SecVolumeSkeleton}

Given $\mathcal{K}$, SkeletonNet further learns a high-resolution skeletal volume $\bm{V} \in \mathbb{R}^{r\times r\times r}$ as the final output, given the ground-truth skeletal volume $\bm{V}^{*} \in \mathbb{R}^{r\times r\times r}$ whose preparation has been described in Sec. \ref{SecSkeletonLearning}. This is efficiently achieved by a proposed module of \emph{globally guided subvolume synthesis}, similar to the scheme used in \cite{han2017high}. The obtained $\bm{V}$ will be subsequently used to recover the surface $\mathcal{M}$ either explicitly or implicitly, as respectively to be presented in Sec. \ref{SecExpMesh} and Sec. \ref{SecImpMesh}.

Specifically, the proposed module is composed of two parallel streams of volume synthesis (cf. Fig. \ref{fig:skeletonNet}). To prepare the module inputs, we first convert $\mathcal{K}$ as a volume $\bm{U}_{in} \in \mathbb{R}^{r'\times r'\times r'}$, with $r' < r$; we then downsample $\bm{U}_{in}$ to have $\bm{U}_{in}^{\downarrow} \in \mathbb{R}^{r'/2\times r'/2\times r'/2}$. To make the conversion from $\mathcal{K}$ to $\bm{U}_{in}$ differentiable, we present a layer of \emph{Point2Voxel} as described shortly in Sec. \ref{SecSkeEnd2End}. The top, global stream learns to refine the input $\bm{U}_{in}^{\downarrow}$ to have the output $\bm{U}_{out}^{\downarrow} \in \mathbb{R}^{r'/2\times r'/2\times r'/2}$, given the ground-truth $\bm{U}^{\downarrow *}$ which is also obtained from $\bm{V}^{*}$; note that $\bm{U}_{out}^{\downarrow}$ has the same resolution as that of $\bm{U}_{in}^{\downarrow}$. The bottom, local stream uniformly samples overlapped subvolumes from $\bm{U}_{in}$, and learns to super-resolve them to have their upsampled versions; for each sampled subvolume $\bm{P}_{in} \in \mathbb{R}^{s' \times s' \times s' }$, the bottom stream learns to generate the upsampled $\bm{P}_{out} \in \mathbb{R}^{s \times s \times s }$, given the ground-truth $\bm{P}^{*}$ sampled from the corresponding 3D position of $\bm{V}^{*}$, and the global $\bm{U}_{out} \in \mathbb{R}^{r\times r\times r}$ (i.e., $\bm{V}$) is obtained by averaging these locally upsampled subvolumes at overlapping voxels. To implement the globally guided subvolume synthesis, we first use an independent auto-encoder (e.g., 3D-R2N2 \cite{choy20163d}) that learns to map the input image $\bm{I}$ as a feature volume of the size $r'/4\times r'/4\times r'/4$, which is used as the augmented input of the top, global stream; we then use the global refinement of $\bm{U}_{out}^{\downarrow}$ as the concatenated input of the bottom, local stream, when it processes each sampled subvolume $\bm{P}_{in}$. These heuristics aim to both correct the possible errors in the predicted $\mathcal{K}$, and improve the consistency among synthesized subvolumes in $\bm{V}$; we empirically find that they are effective for generation of a high-quality $\bm{V}$.

In this work, we implement both the global and local streams based on the 3D U-Net architecture \cite{ronneberger2015u}, whose specifics are given in Sec. \ref{SecExp}. We set $r' = r/2$, $s'=r'/4+4$, and $s = 2s'$, which, when $r = 256$,  gives $|\bm{U}_{in}^{\downarrow}| = |\bm{U}_{out}^{\downarrow}| = 64^3$, $|\bm{P}_{in}| = 36^3$, $|\bm{P}_{out}| = 72^3$, $|\bm{U}_{in}| = 128^3$, and $|\bm{U}_{out}| = |\bm{V}| = 256^3$. To train the module, we use a per-voxel binary cross-entropy loss defined as
\begin{eqnarray}\label{EqnVolumeRefineLoss}
L_{refine} = \sum_{\bm{i} \in \{\bm{j} | \bm{j} \in [1:r]^3\}  } \bm{V}^{*}(\bm{i}) \log \bm{V}(\bm{i}) \ \ + \nonumber \\ \quad\quad\quad\quad\quad\quad\quad\quad\quad (1 - \bm{V}^{*}(\bm{i})) \log (1 - \bm{V}(\bm{i})) .
\end{eqnarray}
In practice, we first train the top, global stream and the bottom, local stream separately, and then train them together in a joint optimization manner.

\vspace{0.1cm}
\noindent \textbf{\emph{Remarks}}
We emphasize that our use of the globally guided subvolume synthesis is both effective and efficient for the generation of high-quality skeletal volumes. Alternative solutions include (1) directly quantizing the skeletal points in $\mathcal{K}$ as voxels, which is, however, not guaranteed to connect the quantized voxels as skeletal curves and skeletal sheets in the 3D volumetric space, and may lose the chance to correct the possible errors in the predicted $\mathcal{K}$; (2) predicting $\bm{V}$ directly from $\mathcal{K}$ via a 3D CNN, which is, however, computationally too expensive to generate a high-resolution $\bm{V}$. We present empirical studies that show the advantages of our used module over these alternatives.

\subsection{End-to-End Training}
\label{SecSkeEnd2End}
The key to make our SkeletonNet an end-to-end trainable system is a differentiable Point2Voxel layer, which converts $\mathcal{K}$ as an input of skeletal volume synthesis.
Without the differentiable Point2Voxel layer, the network for generation of skeletal points is only supervised by two loss terms, i.e. $L_{\phi}(\mathcal{K}_{Cur})$ and $L_{\psi}(\mathcal{K}_{Sur})$; the term $L_{\phi}(\mathcal{K}_{Cur})$ is optimized to enforce the consistency between predicted skeletal curves  $\mathcal{K}_{Cur}$ and its corresponding ground-truth $\mathcal{K}_{Cur}^{*}$, the term $L_{\psi}(\mathcal{K}_{Sur})$ is responsible for constraining the predicted skeletal sheets $\mathcal{K}_{Sur}$ to well approximate $\mathcal{K}_{Sur}^{*}$, and no constraints would be imposed to guarantee the assembly of predicted skeletal curves and sheets, i.e. $\mathcal{K} = \mathcal{K}_{Cur} \bigcup \mathcal{K}_{Sur}$, to approximate $\mathcal{K}^{*}$. The differentiability of the Point2Voxel layer enables that the gradient of the volumetric skeleton refinement network can propagate to the skeletal point generation network, and consequently can constrain the consistent assembly of predicted skeletal curves and sheets.

\vspace{0.1cm}
\noindent \textbf{\emph{A Point2Voxel layer}}
To assign a value to any voxel $\bm{U}_{in}(\bm{i})$, we use the following soft scheme
\begin{equation}\label{EqnPoint2Voxel}
\bm{U}_{in}(\bm{i}) = \exp\left(-M \min_{\bm{p} \in \mathcal{K}} \| \bm{c}(\bm{i}) - \bm{p} \|_2^2 \right) ,
\end{equation}
where $M$ is a scaling constant and $\bm{c}(\bm{i}) \in \mathbb{R}^3$ denotes coordinates of the voxel center. Consider a large scaling $M$; for voxels whose spatial locations are closest to any skeletal points in $\mathcal{K}$, the scheme (\ref{EqnPoint2Voxel}) assigns values approaching to $1$, and for those far away, the assigned values decay rapidly to $0$. To improve the practical efficiency, we only consider those voxels whose center points fall in a small neighborhood of any $\bm{p} \in \mathcal{K}$, while directly nulling other voxels.

Combining the loss terms (\ref{EqnCurSkeDecoderLoss}), (\ref{EqnSurSkeDecoderLoss}), and (\ref{EqnVolumeRefineLoss}) gives the overall training objective of our proposed SkeletonNet
\begin{align}\label{EqnSkeletonNetTrainObj}
L_{SkeNet} = L_{\phi}(\mathcal{K}_{Cur}) + L_{\psi}(\mathcal{K}_{Sur}) + \beta L_{refine}(\bm{V}; \mathcal{K}) ,
\end{align}
where $\beta$ is a balancing parameter.

\section{An Explicit Learning of Mesh Recovery from Skeleton}
\label{SecExpMesh}
 \begin{figure}[h]
    \begin{center}
	\includegraphics[scale=0.3]{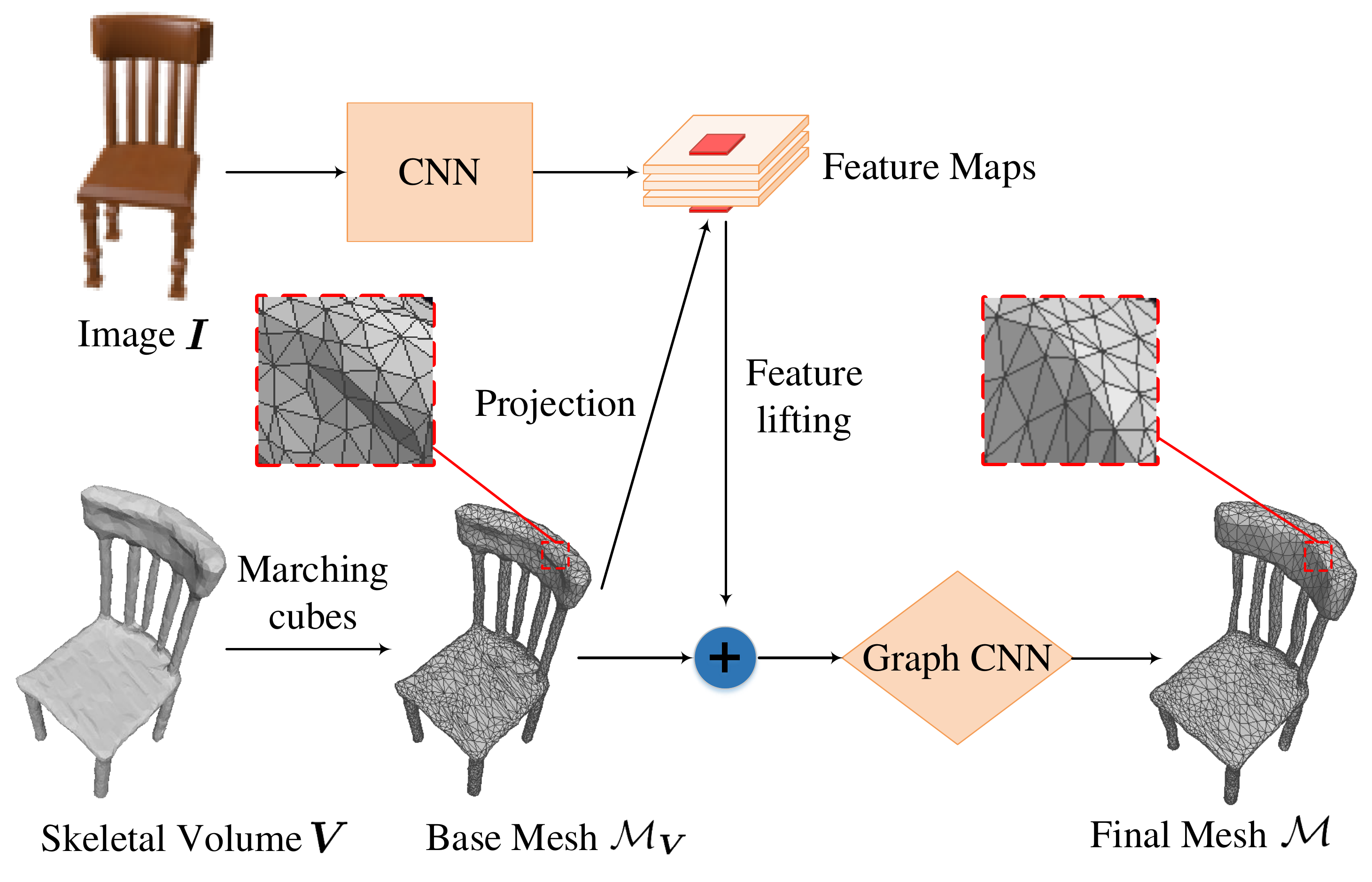}\\
	\caption{The pipeline of our proposed Skeleton-Based Graph CNN (SkeGCNN) for explicit mesh recovery from an input image.} 
	\label{fig:mesh_refine_alg}
    \end{center}
\end{figure}

In this section, we study how to explicitly learn a recovery of surface mesh from the skeletal volume $\bm{V}$ produced by SkeletonNet. This is possible considering the property that a surface mesh and its corresponding skeleton share the same topological structure. Technically, we consider extraction of an initial mesh $\mathcal{M}_{\bm{V}}$ from $\bm{V}$, and then learning to deform the vertices of $\mathcal{M}_{\bm{V}}$ such that $\mathcal{M}_{\bm{V}}$ is inflated to fit the object surface. We present specific steps of the process as follows.

\vspace{0.1cm}
\noindent\textbf{\emph{Extraction of the Initial Mesh}}
We use marching cubes ~\cite{lorensen1987marching} to produce the initial $\mathcal{M}_{\bm{V}}$ from $\bm{V}$. Since the volume $\bm{V}$ is in high resolution, $\mathcal{M}_{\bm{V}}$ would contain a large number of vertices and faces. This makes the subsequent mesh inflation possible to fit an object surface that has complex surface geometries.

\vspace{0.1cm}
\noindent\textbf{\emph{Mesh Deformation using Graph CNNs}}
The mesh $\mathcal{M}_{\bm{V}} = \{\mathcal{T}_{\bm{V}}, \mathcal{E}_{\bm{V}}, \mathcal{F}_{\bm{V}}\}$ itself has a graph data structure; we thus choose to use a graph CNN (GCNN) to learn the deformations of its vertices in $\mathcal{T}_{\bm{V}}$. For any $\bm{t} \in \mathcal{T}_{\bm{V}}$, denote its associated feature vector at layer $l$ of the GCNN as $\bm{f}_{\bm{t}}^{l}$, and $\mathcal{N}(\bm{t})$ be its local neighborhood; we simply compute its features at layer $l+1$ as
\begin{align}\label{EqnGCNNLayerComputation}
\bm{f}_{\bm{t}}^{l+1} = w_{\bm{t}}^{l} \bm{f}_{\bm{t}}^{l} + \sum_{\bm{t}' \in \mathcal{N}(\bm{t})} w_{\bm{t}'}^{l} \bm{f}_{\bm{t}'}^{l} ,
\end{align}
where $w_{\bm{t}}^{l}$ and $\{ w_{\bm{t}'}^{l} \}$ are the network weights. Assuming that the network has $L$ layers, it is trained to output the coordinate offsets $\delta\bm{t} \in \mathbb{R}^3$ for each $\bm{t}$ at the last layer, i.e., $\bm{f}_{\bm{t}}^{L} = \delta\bm{t}$. To enhance the deformation learning, we learn feature maps from the input $\bm{I}$ using image encoder, and lift the learned pixel-wise features to each vertex $\bm{t} \in \mathcal{T}_{\bm{V}}$; we concatenate the lifted features with vertex coordinates to form the input features of GCNN. For feature lifting, we use the camera pose prediction model of DISN ~\cite{xu2019disn} to estimate camera pose, and use the estimated camera pose to project each vertex $\bm{t} \in \mathcal{T}_{\bm{V}}$ to the 2D image domain. We term such a GCNN based approach for deforming vertices of an initial mesh from skeleton input as Skeleton GCNN (SkeGCNN); Fig. \ref{fig:mesh_refine_alg} gives the illustration.

\vspace{0.1cm}
\noindent\textbf{\emph{Loss Functions and Network Training}} Training of SkeGCNN is to learn network weights that drive the deformations $\Delta\mathcal{T}_{\bm{V}} = \{ \delta\bm{t} | \bm{t} \in \mathcal{T}_{\bm{V}} \}$, such that the resulting vertex set $\mathcal{T} = \{ \delta\bm{t} + \bm{t} | \bm{t} \in \mathcal{T}_{\bm{V}} \} $, together with the vertex connections in $\{\mathcal{E}, \mathcal{F}\}$, defines a mesh prediction $\mathcal{M}$ that is close to the ground-truth mesh $\mathcal{M}^{*}$ under a certain measure of distance. Mesh distance is usually approximated as the distance between point sets sampled from the respective meshes. It is thus straightforward to use the Chamfer distance as the measure, similar to (\ref{EqnCurSkeDecoderLossCD}). In this work, we use a weighted version of Chamfer distance in order to emphasize the learning more on surface regions whose geometries have higher complexities; we technically achieve this by assigning larger weights to points sampled from the high-curvature regions on $\mathcal{M}^{*}$, giving rise to
\begin{align}\label{EqnExplicitMeshLossWeightedCD}
\widetilde{D}(\Delta\mathcal{T}_{\bm{V}}; \mathcal{M}, \mathcal{M}^{*})  = \sum_{\bm{q} \in \mathcal{M}} \min_{\bm{q}^{*} \in \mathcal{M}^{*}} \kappa_{\bm{q}^{*}} \| \bm{q}(\Delta\mathcal{T}_{\bm{V}}) - \bm{q}^{*} \|_2^2 \ \ + \nonumber \\  \sum_{\bm{q}^{*} \in \mathcal{M}^{*}} \kappa_{\bm{q}^{*}} \min_{\bm{q} \in \mathcal{M}} \| \bm{q}(\Delta\mathcal{T}_{\bm{V}}) - \bm{q}^{*} \|_2^2 ,
\end{align}
where $\kappa_{\bm{q}^{*}}$ denotes the weight assigned to a point $\bm{q}^{*}$ sampled from $\mathcal{M}^{*}$, and $\bm{q}(\Delta\mathcal{T}_{\bm{V}})$ is a function of vertex deformations. We set $\kappa_{\bm{q}^{*}}$ empirically by thresholding the largest angle variation of normals associated with points in the local neighborhood of $\bm{q}^{*}$. Note that a point $\bm{q} \in \mathcal{M}$ would distribute its supervision signal to the three vertices defining the mesh face where it is sampled. We practically implement (\ref{EqnExplicitMeshLossWeightedCD}) by sampling a fixed number of points respectively from $\mathcal{M}$ and $\mathcal{M}^{*}$. Except for fitting the predicted $\mathcal{M}$ to $\mathcal{M}^{*}$, we also impose regularization on the learning of SkeGCNN, in order for the learned model to generalize to testing instances. The regularization is mainly to promote the smoothness of resulting meshes. We practically follow \cite{wang2018pixel2mesh} and use a first regularizer that constrains the lengths of edges in $\mathcal{E}$, and a second one that constrains the normal variations for points in local neighborhoods (i.e., promotion of small local curvatures). Our regularized SkeGCNN learning objective is as follows
\begin{align}\label{EqnExplicitMeshTrainObj}
L_{SkeGCNN} = \widetilde{D}(\Delta\mathcal{T}_{\bm{V}}; \mathcal{M}, \mathcal{M}^{*}) + \lambda_1 R_{edge}(\Delta\mathcal{T}_{\bm{V}}; \mathcal{E}) + \nonumber \\ \lambda_2 R_{curvature}(\Delta\mathcal{T}_{\bm{V}}; \mathcal{M}) ,
\end{align}
where $\lambda_1$ and $\lambda_2$ are penalty parameters. One may refer to \cite{wang2018pixel2mesh} for definitions of the two regularizers.

\section{Deep Implicit Surface Field Learning with Regularization of Skeleton}
\label{SecImpMesh}

In this section, we present how to use the skeletal volume $\bm{V}$ produced by SkeletonNet to regularize the learning of an implicit surface field. We first introduce the necessary background.

\begin{figure}[h]
    \begin{center}
	\includegraphics[scale=0.30]{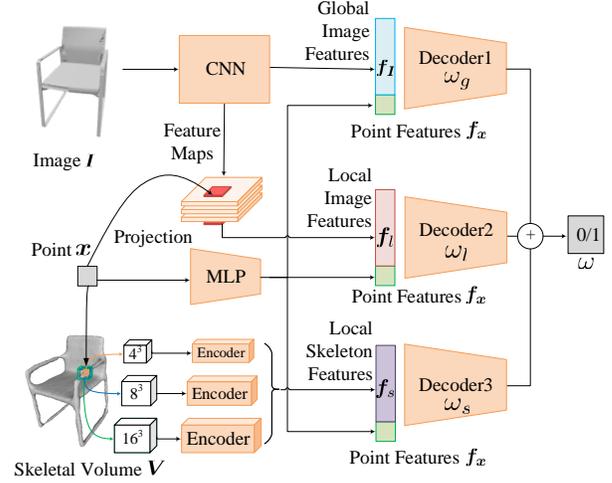}\\
	\caption{An illustration of our proposed Skeleton-Regularized Deep Implicit Surface Network (SkeDISN). }
	\label{fig:imp_mesh_alg}
    \end{center}
\end{figure}

\vspace{0.1cm}
\noindent \textbf{\emph{Deep Implicit Field Learning}}
Except for explicit representation, an object surface can also be represented implicitly as a field function $\omega: \mathbb{R}^3 \rightarrow [0, 1]$ of occupancy probability, which indicates the surface boundary by classifying any $\bm{x} \in \mathbb{R}^3$ as either inside or outside the object, or alternatively as a signed distance function $\omega: \mathbb{R}^3 \rightarrow (-\infty, \infty)$ whose zero-level isosurface is the exact object surface. Recent methods \cite{chen2019learning, mescheder2019occupancy, park2019deepsdf} propose to learn a deep network (typically an MLP) as the implicit function of object surface, such that it can output field values given sampled input points in the 3D space; the obtained field values at sampled 3D points would enable mesh recovery via a final step of marching cubes \cite{lorensen1987marching}. They apply deep implicit field learning to the task setting of RGB-to-mesh recovery, where a CNN based image encoder is typically used to learn latent shape code from the input image, which is used together with coordinates of points in the 3D space as the inputs of MLP based field function. The framework is improved in \cite{xu2019disn} by feeding extra, pixel-wise features extracted from the learned feature maps of the input image.

\vspace{0.1cm}
\noindent \textbf{\emph{Skeleton-Regularized Learning}}
In this work, we are interested in investigating whether the skeletal volume $\bm{V}$ given by SkeletonNet would provide additional benefits to the learning of implicit surface function. To this end, we adopt a network architecture in Fig. \ref{fig:imp_mesh_alg}, which largely follows \cite{xu2019disn} except the skeleton part. We term the proposed model as Skeleton-Regularized Deep Implicit Surface Network (SkeDISN). The network is composed of three parallel streams, which for any input of sampled $\bm{x} \in \mathbb{R}^3$, output 2-dimensional in-out probability vectors of the same format; they also share a same MLP that learns a feature vector $\bm{f}_{\bm{x}}$ from coordinates of the input $\bm{x}$.
The first stream learns a field function $\omega_g: \mathbb{R}^3 \rightarrow [0, 1]^2$ that has an additional input of latent vector $\bm{f}_{\bm{I}}$, encoding the global surface shape of the object in the input image $\bm{I}$. The second stream extracts pixel-wise features $\bm{f}_l$ from feature maps of the image encoder, for which estimated camera pose is used to project the 3D point $\bm{x}$ to the image domain, and concatenates the extracted features with $\bm{f}_{\bm{x}}$ to form the input of a field function $\omega_l: \mathbb{R}^3 \rightarrow [0, 1]^2$. The last stream implements learning from the skeletal volume $\bm{V}$; for the sampled $\bm{x}$, it extracts from $\bm{V}$ multi-scale subvolumes centered at $\bm{x}$, and uses 3D CNNs to learn their respective feature vectors; it concatenates the learned feature vectors to form $\bm{f}_s$, which is then concatenated with $\bm{f}_{\bm{x}}$ to form the input of a field function $\omega_s: \mathbb{R}^3 \rightarrow [0, 1]^2$. The network finally sums up the three probability vectors as $\omega_g(\bm{x}) + \omega_l(\bm{x}) + \omega_s(\bm{x})$, and feeds it into a softmax layer to get the final 0/1 classification. All the three field functions are implemented as MLPs. With a slight abuse of notation, we write the overall field function of occupancy probability implemented by SkeDISN as $\omega: \mathbb{R}^3 \rightarrow [0, 1]^2$.

\vspace{0.1cm}
\noindent \textbf{\emph{Network Training and Inference}}
Given ground-truth object meshes, we use the following objective to train SkeDISN
\begin{equation}\label{EqnSkeDISNTrainObj}
L_{SkeDISN} = \sum_{\bm{x} \in \mathbb{R}^3} \left\{
\begin{array}{ll}
\mathrm{CE}(\omega(\bm{x}), \mathbb{I}_{\mathcal{M}^{*}}(\bm{x})) & \text{if  } d_{\mathcal{M}^{*}}(\bm{x}) < \varepsilon ,\\
0 & \text{otherwise} ,
\end{array} \right.
\end{equation}
where $\text{CE}(\cdot,\cdot)$ stands for a two-way cross-entropy loss, $\mathbb{I}_{\mathcal{M}^{*}}(\bm{x}) \in [0, 1]^2$ is an indicator whose two entries are set as either 1 or 0 depending on whether $\bm{x}$ is inside or outside $\mathcal{M}^{*}$ in the 3D space, and $d_{\mathcal{M}^{*}}(\bm{x})$ represents the distance between $\bm{x}$ and its projection onto $\mathcal{M}^{*}$; by thresholding the distance at a small $\varepsilon$, the loss (\ref{EqnSkeDISNTrainObj}) is activated only when the sampled points are close to the object surface. Given a trained SkeDISN, one would sample points in the 3D space and forward-propagate them through the network to get their in-out classification results; the mesh recovery can be finally obtained via marching cubes~\cite{lorensen1987marching}.

\section{Experiments}
\label{SecExp}

\vspace{0.1cm}
\noindent \textbf{\emph{Dataset}} Our training dataset is mainly from ShapeNet~\cite{chang2015shapenet}, where we collect 43,784 3D shapes in 13 categories. The dataset is split into two parts, $80\%$ shapes are used for training and the remaining for testing. For each 3D model, we rendered 24 RGB images of size $224*224$ under different viewpoints, forming the image-model pairs. The shape categories include: Plane, Bench, Cabinet, Car, Chair, Monitor, Lamp, Speaker, Firearm, Couch, Table, Cellphone, Watercraft. Note that, as all existing approaches, the models in our dataset are aligned in a unified object-centric coordinate frame.\newline

\vspace{0.1cm}
\noindent\textbf{\emph{Model Parameters and Implementation Details}}
For \emph{SkeletonNet}, we use ResNet18~\cite{he2016deep} as the image encoder, which produces $512$-dimensional feature vectors as the latent shape encoding. Both the MLPs of CurSkeDecoder and SurSkeDecoder stack $4$ fully-connected (FC) layers, whose numbers of output neurons are respectively $512$, $256$, $128$ and $3$, where ReLU activation is used for the first $3$ layers and tanh activation is used for the last layer. To generate points of skeletal curves in each $\mathcal{K}_{Cur}$, we use $20$ 1D primitives of the segment $[0, 1]$ as the input of CurSkeDecoder; we also use $20$ 2D primitives of the unit square $[0, 1]^2$ as the input of SurSkeDecoder, for generation of points in each $\mathcal{K}_{Sur}$. We set the penalty parameter in (\ref{EqnCurSkeDecoderLoss}) and  (\ref{EqnSurSkeDecoderLoss}) as $\alpha = 0.2$, and that in (\ref{EqnSkeletonNetTrainObj}) as $\beta = 1$.
Module specifics for volumetric skeleton refinement are as follows. Implementations of both the global and local streams are based on 3D U-Net~\cite{ronneberger2015u}; that for the global stream consists of 4 convolution layers with the respective numbers of channels as 32, 64, 128, and 128, and 4 deconvolution layers with the respective numbers of channels as 128, 64, 32, and 2; for local stream, the 3D U-Net consists of 4 convolution layers with the respective numbers of channels as 32, 64, 128, and 128, and 5 deconvolution layers with the respective numbers of channels as 128, 64, 32, 16, and 2; all the convolution/deconvolution layers are with kernels of size 3 and stride 2. For the layer of Point2Voxel, we set the scaling constant in (\ref{EqnPoint2Voxel}) as $M = 10$. We implement the end-to-end training of SkeletonNet as follows. We first train CurSkeDecoder and SurSkeDecoder separately, for which we use a learning rate of 1e-3 for 150 epochs, and then train the module for volumetric refinement using a learning rate of 1e-4 for 20 epochs; we finally fine-tune the entire SkeletonNet with a learning rate of 1e-5 for 20 epochs.

\noindent Our SkeGCNN for explicit mesh recovery consists of 6 graph convolution layers, each of which learns weights to perform feature learning, according to (\ref{EqnGCNNLayerComputation}). The first layer receives a pixel-wise feature vector of dimension 963; all the hidden layers work with feature vectors of dimension 192; the last layer outputs 3-dimensional coordinate offsets. We use VGG-16~\cite{simonyan2015very} as the image encoder of SkeGCNN, and lift pixel-wise features at 'conv4', 'conv7', 'conv10', and 'conv13' of VGG-16 to vertices of the initial mesh. To set values of $\kappa_{\bm{q}^{*}}$ in (\ref{EqnExplicitMeshLossWeightedCD}), we construct the local neighborhood of size 16 around $\bm{q}^{*}$, and set $\kappa_{\bm{q}^{*}} = 5$ when the largest angle variation associated with points in the neighborhood is beyond $60^\circ$, and set $\kappa_{\bm{q}^{*}} = 1$ otherwise. We set the penalty parameters $\lambda_1 = 7e^{-1}$ and $\lambda_2 = 3e^{-4}$ in the SkeGCNN training objective (\ref{EqnExplicitMeshTrainObj}). SkeGCNN is trained with an initial learning rate of 1e-4, which is decayed by a factor of 10 every 20 epochs until a total of 60 epochs.

\noindent Our SkeDISN uses the same image encoder as in SkeGCNN. The MLP of feature embedding for coordinates of sampled points consists of three FC layers respectively of 64, 128, and 512 neurons. Each of the MLPs implementing the three parallel field functions $\omega_g$, $\omega_l$, and $\omega_s$ consists of three FC layers respectively of 512, 256, and 2 neurons, except for their respective input layers. For use of features in the input skeletal volume, we extract its respective subvolumes of sizes $4^3$, $8^3$, and $16^3$ centered at each sampled point. We set the threshold $\varepsilon = 0.1$ in the SkeDISN training  objective (\ref{EqnSkeDISNTrainObj}). SkeDISN is trained with an initial learning rate of 1e-4, which decays at a factor of 0.9 every 2 epochs until a total of 60 training epochs.

\vspace{0.1cm}
\noindent \textbf{\emph{Evaluation Metrics}} We use Chamfer distance (CD) and intersection-over-union (IoU) as the primary metrics for evaluation of mesh recoveries from RGB images. For computation of CD, we do mesh2point conversion by uniformly sampling 10,000 points on each surface mesh. For computation of IoU, we follow~\cite{xu2019disn} and do voxelization in a grid of $64^3$. To evaluate results of skeletal point sets and skeletal volumes given by SkeletonNet, we use CD and IoU as the metrics as well.

\subsection{Evaluation of the Proposed SkeletonNet}
\label{ExpSkeletonNet}
In this section, we evaluate the efficacy of our proposed SkeletonNet for learning skeletal point sets and skeletal volumes from the input RGB images, by comparing with alternative designs. We intensively use the ShapeNet category of \emph{chair} for these comparisons, and give the summarized results on other categories.

\renewcommand\arraystretch{1.2}
    \begin{table}[h]
        	\begin{center}
        		\begin{tabular}{c | c  c}\hline
        			\centering
        			Variant method & CD ($\times 0.001$) & IoU ($\times 100$) \\
        			\hline
        			Point-wise fitting  & 2.218 & 28.97  \\
        			Line-wise fitting    & 2.242 & 35.48  \\
        			Square-wise fitting  & 2.016 & 32.81  \\
        			Shared line-and-square fitting  & 2.451 & 33.10 \\
                    \hline
        			SkeletonNet w/o Laplacian regu.  & 1.911  & 31.10 \\
                    SkeletonNet w/o end-to-end training  & 1.601  & 37.73 \\
        			SkeletonNet  & \textbf{1.485}  & \textbf{39.50}  \\
        			\hline
        		\end{tabular}
        		\caption{Quantitative comparisons between the module of SkeletonNet and its variants for producing the intermediate results of skeletal point set from input RGB images. Results are obtained on the ShapeNet \emph{chair} category. Please refer to the main text for specific settings of these variants. }
        		\label{tab:ske_metric}
        	\end{center}
    \end{table}

    \begin{figure*}[h]
    	\begin{center}
    		\includegraphics[width=0.8\linewidth]{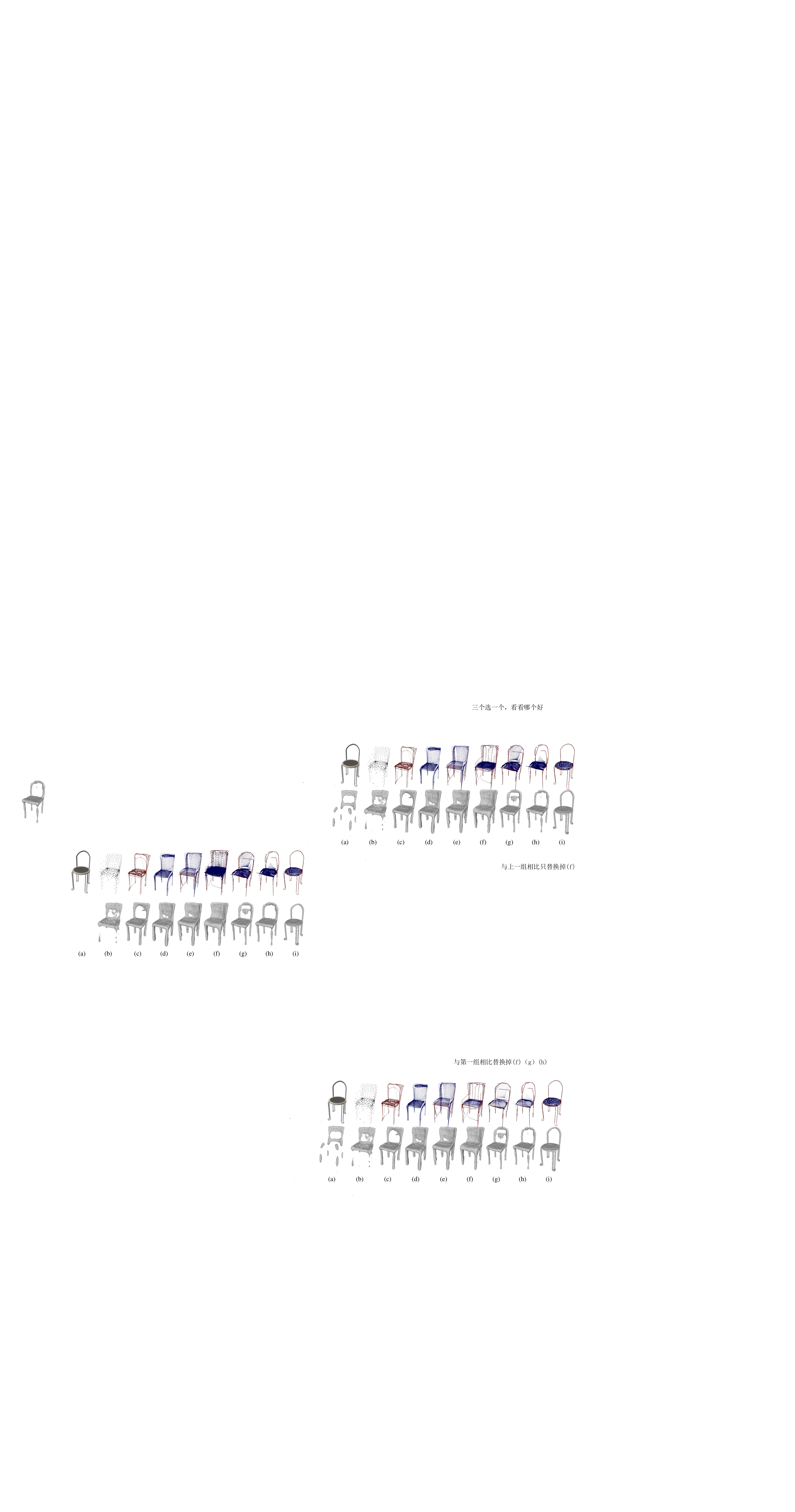} \\
    		\begin{tabular}{p{35pt}p{35pt}p{35pt}p{35pt}p{35pt}p{35pt}p{35pt}p{35pt}p{35pt}}
                    \quad  (a)  & 
                    \quad \quad (b)  &
                    \quad \quad  (c)  & 
                    \quad \ (d) & 
                    \quad \ (e) &
                    \quad  (f) &
                    \quad  (g) &
                     \ (h) &
                     \ (i)\quad
            \end{tabular}
    			\caption{Qualitative comparisons between SkeletonNet and its variants. These variants replace the module of SkeletonNet for producing the intermediate results of skeletal point set from input RGB images. (a) Input image; (b) point-wise fitting; (c) line-wise fitting (with Laplacian regularization); (d) square-wise fitting (with Laplacian regularization); (e) shared line-and-square fitting (with Laplacian regularization); (f) SkeletonNet without Laplacian regularization; (g) SkeletonNet without end-to-end training; (h) SkeletonNet; (i) Ground truth. Please refer to the main text for specific settings of these variants. }
    		\label{fig:com_ske}
    	\end{center}
    \end{figure*}

    \renewcommand\arraystretch{1.2}
    \begin{table}[h]
        	\begin{center}
        		\begin{tabular}{c | c}\hline
        			\centering
        			Variant method &  IoU ($\times 100$) \\
        			\hline
        		    Quantization and morphological dilation  & 27.52 \\
        		    SkeletonNet via subvolume systhesis alone  & 37.64 \\
        			SkeletonNet via globally guided subvol. synthesis  & \textbf{39.50} \\
        			\hline
        		\end{tabular}
        		\caption{Quantitative comparisons between the SkeletonNet module of volumetric refinement and its variants. Results are obtained on ShapeNet \emph{chair} category. Please refer to the main text for specific settings of these variants.}
        		\label{tab:local_global_comp}
        	\end{center}
        	\vspace{-10pt}
    \end{table}
    
    \begin{figure}[h]
    	\begin{center}
    		\includegraphics[scale=0.55]{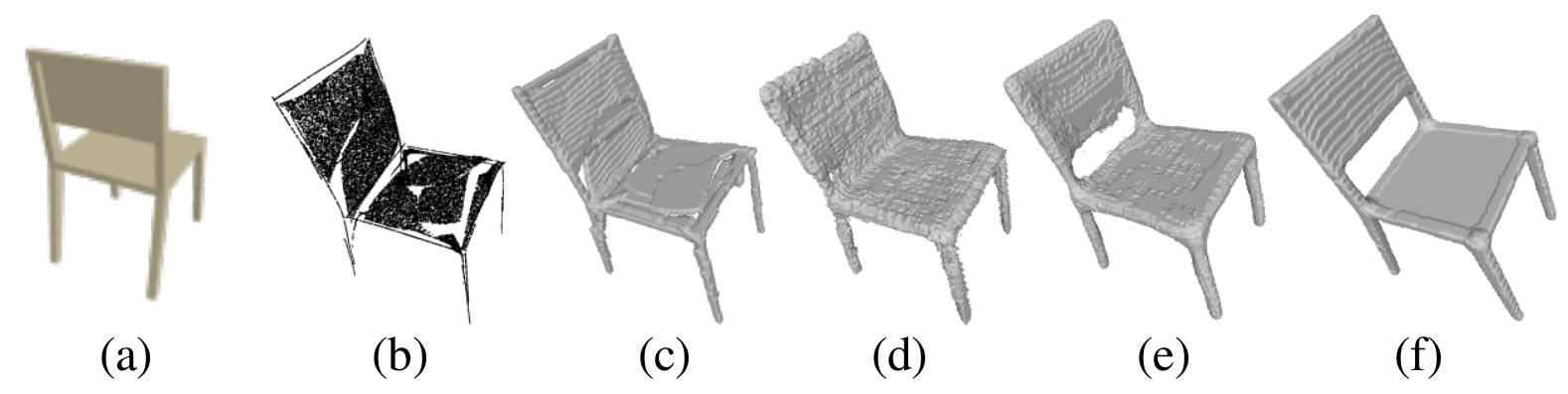}\\
    		\caption{Qualitative comparisons between the SkeletonNet module of volumetric refinement and its variants. (a) Input image; (b) the intermediate result of skeletal point set; (c) quantization and morphological dilation; (d) subvolume synthesis alone; (e) our used module of globally guided subvolume synthesis; (f) ground truth.}
    		\label{fig:local_global_comp}
    	\end{center}
    	\vspace{-15pt}
    \end{figure}

We first evaluate the SkeletonNet module that produces an intermediate result of skeletal point set $\mathcal{K}$ from an input image. The module uses CurSkeDecoder and SurSkeDecoder that respectively regress points on skeletal curves and skeletal sheets, where a regularizer of Laplacian smoothness is used to preserve the neighboring relations of points connected on the input 1D primitives of line or 2D primitives of square. The efficacy of such a design is verified by comparing with 1) ``point-wise fitting'' that directly adopts PSG~\cite{fan2017point} to regress the skeletal points, 2) ``line-wise fitting'' that removes the SurSkeDecoder and only deforms the input 1D primitives for regression of skeletal points, 3) ``square-wise fitting'' that removes the CurSkeDecoder and only deforms the input 2D primitives for regression of skeletal points, and 4) ``shared line-and-square fitting'' that uses a single MLP, instead of using the parallel CurSkeDecoder and SurSkeDecoder, to learn deformations of 1D lines and 2D squares. Note that end-to-end training is enabled in the above variants via our proposed point2voxel layer, and the Laplacian regularizer is also used in the last three of the above variants; to investigate the efficacy by their owns, we further compare with 5) SkeletonNet whose training is not regularized by Laplacian smoothness, and 6) SkeletonNet whose first module for the intermediate skeletal point set and second module of volumetric refinement are separately trained. We make the comparisons based on two measures: 1) CD between the predicted $\mathcal{K}$ and the ground-truth $\mathcal{K}^{*}$, and 2) IoU between the final output $\bm{V}$ from each of these variants and the ground-truth $\bm{V}^{*}$, where the same SkeletonNet module of volumetric refinement is used for producing $\bm{V}$ from $\mathcal{K}$. Table \ref{tab:ske_metric} gives the comparative results, which are obtained on the ShapeNet \emph{chair} category. Clearly, point-, line-, and square-wise deformations may not well fit the skeletal point sets, and shared line-and-square fitting is inferior to be responsible for regression of both skeletal curves and skeletal sheets. It is also observed in Table \ref{tab:ske_metric} that Laplacian regularization is important to constrain the deformations by connecting neighboring points on lines and squares, and gives a high-quality final result of skeletal volume. The end-to-end training enabled by Point2Voxel layer gives rise to more accurate predictions of skeletal point cloud and volume. These observations are further demonstrated by examples of qualitative results in Fig. \ref{fig:com_ske}.

\renewcommand\arraystretch{1.2}
     \begin{table*}[t]
    	\begin{center}
    	\begin{tabular}{c|*{7}{c}}\hline
    		    \centering
    			category & plane & bench & cabinet & car & chair & monitor & lamp \\
    			\hline
    			CD ($\times 0.001$)  & 1.153 & 1.245 & 1.901 & 0.918 & 1.473 & 1.879 & 3.357 \\
    			IoU ($\times 100$) & 38.49 & 36.72 & 55.00 & 67.68 & 38.78 & 35.82 & 35.12 \\
    			\hline
    			category &  speaker & firearm & couch & table & cellphone & watercraft & mean \\
    			\hline
    			CD ($\times 0.001$) & 2.787 & 0.882 & 1.608 & 1.728 & 1.124 & 1.530 & 1.660 \\
    			IoU ($\times 100$) & 54.35 & 38.91  & 47.38 & 40.02 & 55.94 & 44.55 & 45.29\\
    			\hline
    		\end{tabular}
    		\caption{ Quantitative results of our SkeletonNet and its intermediate predictions of skeletal point set on all of 13 categories in ShapeNet\cite{chang2015shapenet}. Skeletal point set is measured by Chamfer distance (CD), and skeletal volume is measured by intersection-over-union (IoU).}
    		\label{tab:ske_allcats}
    	\end{center}
    	\vspace{-10pt}
    \end{table*}

    \begin{figure*}[t]
        	\begin{center}
        	\includegraphics[scale=0.50]{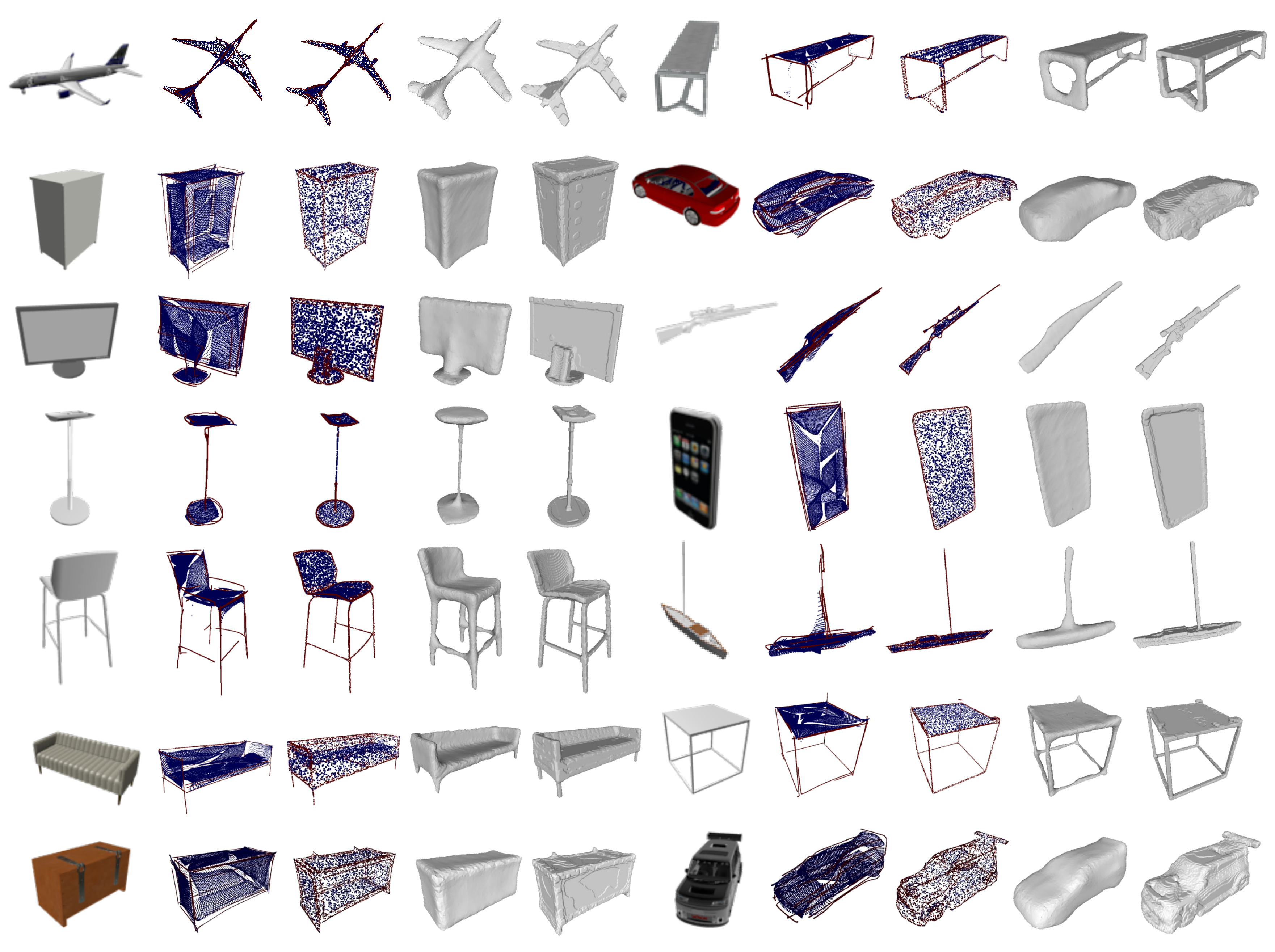}\\
        		\begin{tabular}{p{35pt}p{35pt}p{35pt}p{35pt}p{35pt}p{35pt}p{35pt}p{35pt}p{35pt}p{35pt}}
                    \ (a)  & 
                    \quad \ (b)  &
                    \quad \ \ (c)  & 
                    \quad \ \ (d) & 
                    \quad \  (e) &
                    \quad \quad (a)  & 
                    \quad \ \ \ (b)  &
                    \quad \quad (c)  & 
                    \quad \quad (d) & 
                    \quad \quad \  (e)
            \end{tabular}
        		\caption{Visualization results of our proposed SkeletonNet and its intermediate predictions of skeletal point set on ShapeNet~\cite{chang2015shapenet} dataset. (a) Input images; (b) The produced skeletal points; (c) The ground-truth skeletal points; (d) The refined skeletal volumes; (e) The ground-truth skeletal volumes.}
        		\label{fig:allcats_ske}
        	\end{center}
        	\vspace{-15pt}
    \end{figure*}
\begin{table*}[t]
	\renewcommand\arraystretch{1.2}
	\begin{center}
		\begin{tabular}{*{11}{c}}
			\toprule
			\multirow{2}*{Category} & \multicolumn{5}{c}{Implicit}   & \multicolumn{4}{c}{Explicit} \\
			\cmidrule(lr){2-6} \cmidrule(lr){7-10}
			& OGN &  IMNet & OccNet  & DISN & SkeDISN & P2M & AtlasNet  &TMNet & SkeGCNN\\
			\midrule
			\midrule
           plane   & 4.765  & 1.929  & 1.967 & 1.674 & \textbf{1.243}
                   & 2.816  & 1.733 & 1.459  & \textbf{0.771} \\
           
           bench   & 7.645 & 2.925  & 2.152 & 1.794 & \textbf{0.944}
                   & 2.809  & 2.427  & 2.020  & \textbf{1.037} \\
                   
           cabinet & 2.531 & 2.705 & \textbf{1.773} & 2.177 & 1.921
                   & 2.463 & 2.362   & 1.857 & \textbf{1.468}
                   \\
                   
           car  & 1.129 & 1.704  & 1.354 & 0.962 & \textbf{0.925}
                & 2.880 & 3.426 & 1.692 & \textbf{0.675}
                \\
                
           chair & 6.418 & 2.718 & 2.258 & 1.538 &  \textbf{1.248}   
                 & 2.763 & 1.622 & 1.485 & \textbf{1.138 }
                 \\
                 
           monitor & 3.726 & 2.518 & 2.080 & 2.150 & \textbf{1.084}
                   & 3.027 & 2.005 & 2.637 & \textbf{1.316}
                   \\
                   
           lamp & 11.274 & 8.156 &15.899  & 6.144 & \textbf{3.958}   
                & 5.357 & 4.609 & 5.450 & \textbf{2.540}
                   \\
                   
           speaker  & 5.772 & 4.583 & 3.392 & 3.327 & \textbf{2.582}   
                 & 7.142 & 3.173 & 3.486 & \textbf{2.446}
                   \\
                   
           firearm & 1.935 & 2.259  &2.128  & 1.310 & \textbf{0.645}
                   & 1.016 & 2.754 & 1.706 & \textbf{0.685}
                   \\
                   
           couch   & 4.488 & 2.574 & 1.755 & 1.850  & \textbf{1.141}   
                   & 2.482 & 2.038 & \textbf{1.049} & 1.256
                   \\
        
           table & 7.178 & 4.239 & 2.739 & 3.127 & \textbf{1.917}
                 & 5.470 & 2.246 & 2.540 & \textbf{1.718}
                   \\
           
           cellphone  & 2.137 & 1.322  &1.366  & 1.036 & \textbf{0.839}
                   & 1.487  &1.699  & \textbf{1.088}  & 1.127
                   \\
                   
           watercraft & 3.941 & 3.819 & 3.319 & 3.165 & \textbf{1.885}
                   & 3.049 & 2.428 & 2.318 & \textbf{1.064}
                   \\
        \midrule
            mean    & 4.842 & 3.189 & 3.249  & 2.327 & \textbf{1.564} 
	        & 3.263  & 2.502   & 2.214 & \textbf{1.326}\\
        \bottomrule
        \end{tabular}
        \caption{Quantitative comparisons (Chamfer distance $\times 0.001$) of our method against state-of-the-arts on ShapeNet~\cite{chang2015shapenet} dataset. The lower, the better. }
        \label{Tab:cd}
        \end{center}
\end{table*}
\begin{table*}[t]
    
	\renewcommand\arraystretch{1.2}
	\begin{center}
		\begin{tabular}{*{11}{c}}
			\toprule
			\multirow{2}*{Category} & \multicolumn{5}{c}{Implicit}   & \multicolumn{4}{c}{Explicit} \\
			\cmidrule(lr){2-6} \cmidrule(lr){7-10}
			& OGN &  IMNet & OccNet  & DISN  & SkeDISN & P2M & AtlasNet & TMNet & SkeGCNN \\
			\midrule
			\midrule
           plane   & 50.44 & 58.33 & 53.45 & 65.01 & \textbf{66.41}
                   & 31.43  & 53.73 & 59.06 & \textbf{61.06} \\
           
           bench   & 48.85 & 56.54 & 56.81 & 61.61 & \textbf{64.30}
                   & 42.11 & 48.79 & 53.42 & \textbf{58.31} \\
                   
           cabinet & 61.98 & 57.63 & 61.92 & 64.45 & \textbf{67.05}
                   & 58.43 & 41.23 & 52.31 & \textbf{59.22} \\
                   
           car     & 81.50 & 78.57 & 79.22  & 83.56 & \textbf{84.36}
                   & 55.25 & 51.15 & 72.38  & \textbf{76.03} \\
                
           chair   & 54.70 & 55.14 & 61.10 & 60.54 & \textbf{64.97}
                   & 50.70 & 48.56  & 51.02  & \textbf{61.71} \\
                 
           monitor & 55.74 & 61.48 & 57.04 & 63.53 & \textbf{67.37}
                   & 45.75 & 49.88  & 56.75  & \textbf{57.07} \\
                   
           lamp    & 44.15  & 41.73 & 50.17 & 48.32 & \textbf{53.39}
                   & 41.83 & 36.49  & 40.73  & \textbf{52.24} \\
                   
           speaker & 56.90 & 61.29 & 61.99 & 65.34 & \textbf{70.00}
                   & 54.17 & 43.37  & 54.74  & \textbf{60.25} \\
                   
           firearm & 59.08 & 59.88 & 53.96 & 71.86 & \textbf{73.46}
                   & 56.24 & 56.07  & 60.16  & \textbf{65.76} \\
                   
           couch   & 63.95 & 66.13 & 71.09  & 70.88 & \textbf{74.72}
                   & 59.70 & 49.17 & 58.76  & \textbf{63.13} \\
        
           table   & 55.36 & 50.77 & 59.78 & 55.44 & \textbf{61.62}
                   & 40.99 & 46.52 & 47.92  & \textbf{58.35} \\
           
           cellphone  & 69.59 & 73.44 & 70.96  & \textbf{76.11} & 75.55
                      & 69.36 & 53.60 & \textbf{72.18}  & 69.43 \\
                   
           watercraft & 55.95 & 61.78 & 58.47  & 63.14 & \textbf{69.09}
                      & 44.52 & 56.12  & 58.44  & \textbf{63.31} \\
        \midrule
            mean     & 58.32 & 60.21 & 61.23 & 65.34 & \textbf{68.64}
                     & 50.04  & 48.82 & 56.68  & \textbf{62.46} \\
        \bottomrule
        \end{tabular}
        \caption{Quantitative comparisons (Intersection over Union $\times 100$) of our method against state-of-the-arts on ShapeNet~\cite{chang2015shapenet} dataset. The higher, the better.}
        \label{Tab:iou}
        \end{center}
        \vspace{-15pt}
\end{table*}

\noindent\begin{figure*}[t]
    \begin{center}
    \includegraphics[scale=0.44]{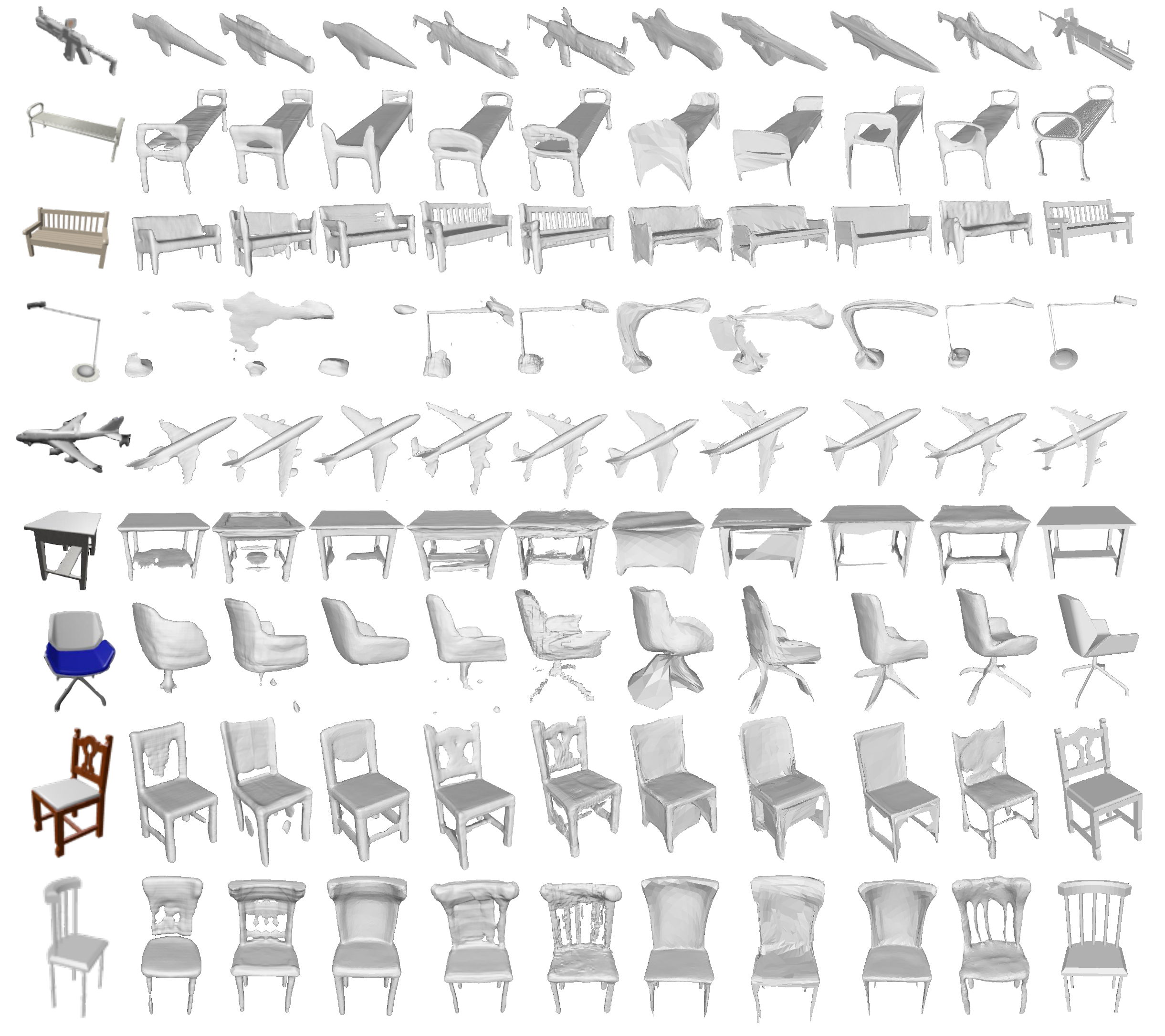}\\
    \begin{tabular}{p{35pt}p{35pt}p{35pt}p{35pt}p{35pt}
    p{35pt}p{35pt}p{35pt}p{35pt}p{35pt}p{35pt}}
                    \quad \quad  (a)  &
                    \quad \quad (b)  &
                    \quad \ (c)  &
                    \quad \ (d) &
                    \quad \ \ (e) &
                    \quad \ (f) &
                    \quad   (g) &
                    \quad  (h) &
                    \quad \ (i) &
                    \quad \ (j) &
                    \  (k)
            \end{tabular}
    \end{center}
    \caption{ (a) Input Images (b) OGN;  (c) IMNet; (d) OccNet;  (e) DISN;  (f) SkeDISN; (g) Pixel2Mesh; (h) AtlasNet; (i) TMNet; (j) SkeGCNN; (k) Ground Truths.}
    \label{fig:comp_recon}
    \vspace{-10pt}
\end{figure*}

We further evaluate the SkeletonNet module of volumetric refinement, which produces the final output $\bm{V}$ from the intermediate $\mathcal{K}$. Given $\mathcal{K}$, $\bm{V}$ can be alternatively obtained by 1)  directly quantizing skeletal points in $\mathcal{K}$ into voxels and performing morphological dilation, and 2) using subvolume synthesis of the module without global guidance. Quantitative results in Table. \ref{tab:local_global_comp} confirm the efficacy of our used module. Examples of qualitative comparisons are shown in Fig. \ref{fig:local_global_comp}, where we observe that the global guidance is important to recover the topological structure that may have been lost in the intermediate result of skeletal point set.

We finally report in Table \ref{tab:ske_allcats} our results of skeletal point set and skeletal volume on other categories of ShapeNet. These results tell that the difficulties of learning skeletons for different categories vary greatly, with cabinet, cellphone, and couch as the easier ones; this is reasonable since most of the instances from these categories have simple topologies. Example results are also shown in Fig. \ref{fig:allcats_ske}; for shapes with simple topologies such as cars and cabinets, the calculated ground-truth skeletal points are usually not compact. This is due to the difficulty of sinking sampled surface points into the center of the objects. However, this does not affect the operation of our algorithm for those simple shapes. Complex topologies and thin structures, such as rods of chair, legs of table, and holder of lamp, are successfully recovered in the resulting skeletons, which are expected to help the downstream task of mesh recovery.

\subsection{Explicit and Implicit Mesh Recoveries from Skeletons}
\label{ExpMeshRecon}
Skeleton is an interior representation of object surface that preserves the surface topology; in this section, we conduct experiments to study its benefits to recovery of surface mesh. The studies are conducted under the task setting of learning mesh recovery from a single RGB image. Specifically, for explicit mesh recovery, we use the skeletal volume produced by SkeletonNet as input of our SkeGCNN model introduced in Sec. \ref{SecExpMesh}; for learning an implicit surface field, we use the SkeDISN introduced in Sec. \ref{SecImpMesh}, which extracts features from the skeletal volume to regularize the field learning. We compare SkeGCNN and SkeDISN with the corresponding groups of state-of-the-art methods.

\begin{figure*}[h]
        \centering
        \includegraphics[scale=0.28]{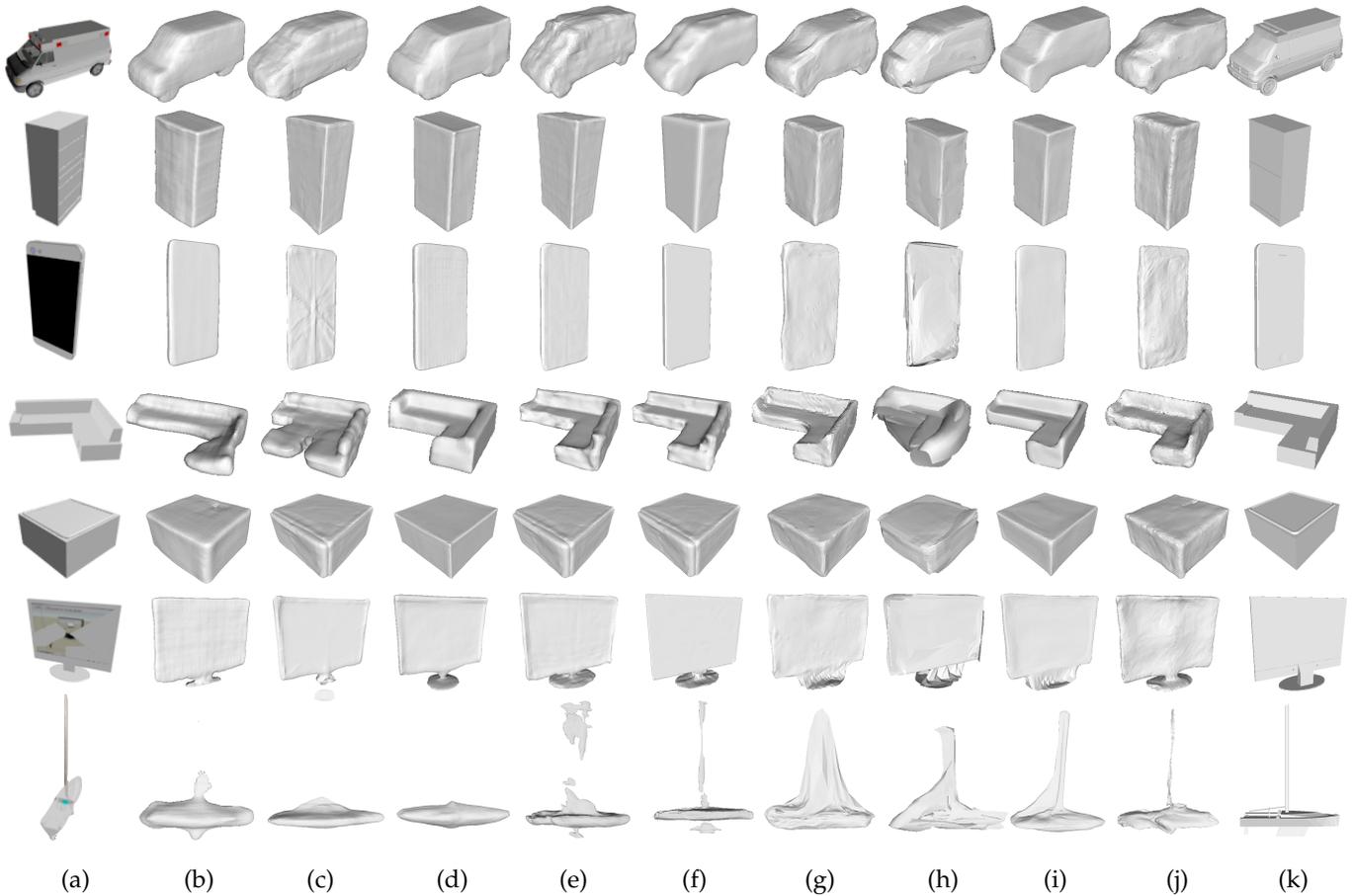}\\
        \begin{tabular}{p{35pt}p{35pt}p{35pt}p{35pt}p{35pt}
        p{35pt}p{35pt}p{35pt}p{35pt}p{35pt}p{35pt}}
                        \quad \quad  (a)  &
                        \quad \quad (b)  &
                        \quad \quad (c)  &
                        \quad \quad \ (d) &
                        \quad \quad \ (e) &
                        \quad \quad \ (f) &
                        \quad \quad \  (g) &
                        \quad \quad \ (h) &
                        \quad \quad (i) &
                        \quad \quad (j) &
                        \quad \ \  (k)
                \end{tabular}
        \caption{ (a) Input Images (b) OGN;  (c) IMNet; (d) OccNet;  (e) DISN;  (f) SkeDISN; (g) Pixel2Mesh; (h) AtlasNet; (i) TMNet; (j) SkeGCNN; (k) Ground Truths.}
        \label{fig:appendix:comp_recon}
        \vspace{-10pt}
    \end{figure*}

Learning a surface mesh explicitly via vertex deformation is pioneered by AtlasNet~\cite{groueix2018atlasnet}; due to its use of multiple initial meshes, the method fails to produce a closed, watertight mesh. Pix2Mesh~\cite{wang2018pixel2mesh} resolves this issue by deforming a single mesh of fixed topology. TMNet~\cite{pan2019deep} further improves the technique to generate object surface of complex topology, by designing a module of face pruning that is able to dynamically adjust the topology during the deformation process. We quantitatively compare our SkeGCNN with these methods in Table~\ref{Tab:cd} under the measure of CD and Table~\ref{Tab:iou} under the measure of IoU,  where results are obtained by  13 ShapeNet categories. We use the publicly released codes to implement these methods; for a fair comparison, their training hyper-parameters have been optimally tuned to the data used in our experiments. Under both the measures of CD and IoU, our results of deforming the input skeletons are significantly better than those of existing methods on most of the 13 categories, except the categories of \emph{couch} and \emph{cellphone}, whose topologies are relatively simple and for which TMNet achieves the best results.
Qualitative results in Fig.~\ref{fig:appendix:comp_recon} demonstrate that our method achieves comparable performance on those objects of simple topologies.
Examples of qualitative results in Fig.~\ref{fig:comp_recon} tell that in terms of generating meshes of complex topologies, our method and TMNet are able to generate mesh results with some thin structures, and our results tend to be better than those of TMNet, while AtlasNet and Pix2Mesh fail to do so.

Mesh recovery via implicit field learning gains an increased recent popularity, with methods proposed for learning deep field functions whose outputs are in different formats. For example, IMNet~\cite{chen2019learning} and OccNet~\cite{mescheder2019occupancy} learn shape-decoding networks whose binary outputs indicate whether a sampled point in the 3D space is inside or outside the object surface; DISN~\cite{xu2019disn} learns a network of signed distance function whose output represents the distance of the sampled point to the underlying surface. Our SkeDISN largely follows DISN; we adapt its architecture a bit to output binary results of in-out classification, which is more compatible to include features from the input skeletal volume. As a baseline, we also compare with OGN~\cite{hane2017hierarchical}, which is the state-of-the-art method to output octree-based surface volume.  Again, these methods are implemented using their publicly released codes, with hyper-parameters optimally tuned for the data of our experiments. Quantitative results in Tables~\ref{Tab:cd} and~\ref{Tab:iou} confirm the efficacy of using skeleton features to regularize the implicit field learning, where results of our SkeDISN are significantly better than those of existing methods under both the measures of CD and IoU.
Qualitative results in Fig.~\ref{fig:appendix:comp_recon} illustrate that our SkeDISN achieves comparable performance on those simple objects. Qualitative results in Fig.~\ref{fig:comp_recon} suggest that DISN and our skeleton-regularized version are the better methods to reconstruct object meshes of complex topologies, since both DISN and SkeDISN utilize the lifted image features. Due to the lack of explicit global topology understanding, DISN tends to fail for recovering long and thin parts; in contrast, our SkeDISN can better preserve these structures.

By comparing the results in Tables~\ref{Tab:cd} and~\ref{Tab:iou} between explicit and implicit methods, one may observe that there exists no a generally good choice of current measure to quantitatively evaluate different methods from all perspectives; CD may better evaluate the algebraic quality of recovered meshes, for which explicit methods perform better, and IoU may better evaluate the topological quality of recovered meshes, for which implicit methods seem prevail. However, visual comparisons in Fig.~\ref{fig:comp_recon} suggest that explicit methods still have their merits even in terms of recovering object meshes of complex topologies. These results suggest that further studies in future research are necessary to understand the respective pros and cons of explicit and implicit surface recovery methods.

    \begin{table}[h]
        	\begin{center}
        		\begin{tabular}{c | c  c}\hline
        			\centering
        			Method & CD ($\times 0.001$) & IoU ($\times 100$) \\
        			\hline
        			OGN + SkeGCNN & 4.485  & 49.84 \\
        			SkeletonNet + SkeGCNN   & \textbf{1.154}  & \textbf{62.90} \\
        		    \hline
        			OGN + SkeDISN  &  1.534  & 61.70  \\
        			SkeletonNet + SkeDISN & \textbf{1.230}  & \textbf{65.20} \\
        			\hline
        		\end{tabular}
        		\caption{Ablation studies on the effectiveness of SkeletonNet for the downstream tasks of explicit and implicit mesh recoveries. Results are obtained on the ShapeNet \emph{chair} category. }
        	    \label{tab:wo_skenet}
        	\end{center}
        	\vspace{-15pt}
    \end{table}

    \begin{figure}[h]
        \vspace{-10pt}
    	\begin{center}
    		\includegraphics[scale=0.375]{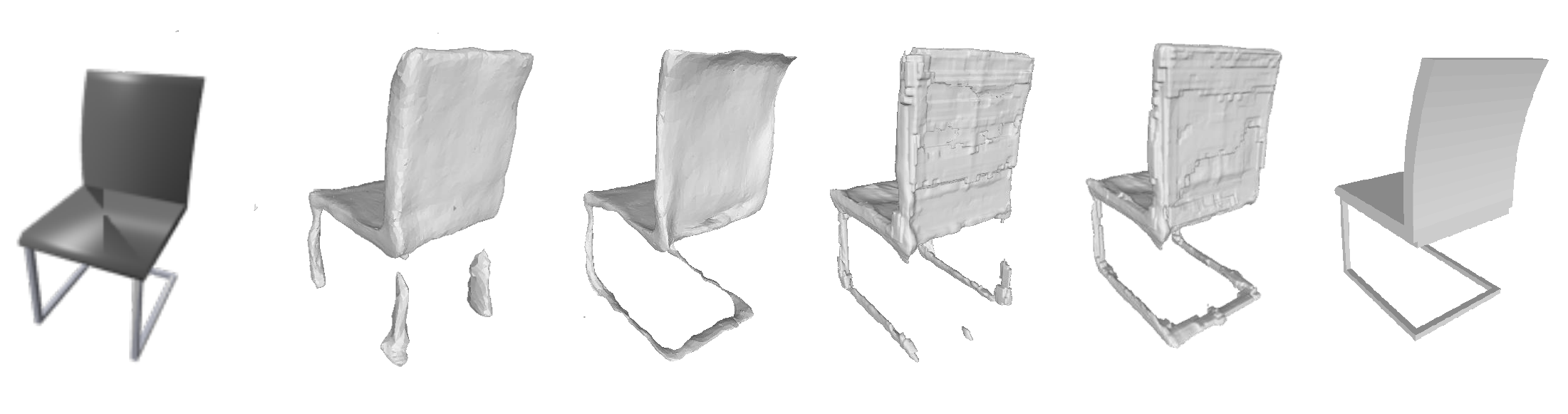}\\
    		 \begin{tabular}{p{30pt}p{30pt}p{30pt}p{30pt}p{30pt}p{30pt}}
                     \quad (a)  & 
                     \quad \ (b)  &
                     \quad \ (c)  & 
                     \quad \ (d) & 
                     \quad \ (e) &
                     \ (f) 
            \end{tabular}
    		\caption{ Example results of the ablation studies on the effectiveness of SkeletonNet for the downstream tasks of explicit and implicit mesh recoveries. (a) Input image; (b) OGN + SkeGCNN; (c) SkeletonNet + SkeGCNN;  (d) OGN + SkeDISN; (e) SkeletonNet + SkeDISN; (f) Ground truth.}
    		\label{fig:wo_skenet}
    	\end{center}
    	\vspace{-10pt}
    \end{figure}
\vspace{0.1cm}
\noindent \textbf{\emph{Ablation studies}}
    Our proposed SkeGCNN and SkeDISN use as inputs the skeletal volumes produced by SkeletonNet. In fact, skeletal volumes may also be obtained by training a state-of-the-art volume synthesis method, e.g., OGN~\cite{tatarchenko2017octree}, using the ground-truth skeletal volumes in our ShapeNet-Skeleton dataset. We thus conduct ablation studies by replacing SkeletonNet with an OGN based volume synthesis network. We note that OGN is considered as an efficient method for synthesizing high-resolution volumes (e.g., volumes of $256^3$). We conduct the ablation experiments on the ShapeNet category of \emph{chair}. Quantitative results in Table~\ref{tab:wo_skenet} confirm that for both explicit and implicit mesh recoveries, our proposed SkeletonNet provides skeletal volumes that are more effective for the downstream tasks. Our advantage can be attributed to the learning of the intermediate skeletal point sets. Example results in Fig.~\ref{fig:wo_skenet} reflect this advantage.

\subsection{Additional Results}
\label{ExpAddRes}

In this section, we present additional results, mostly in variant task settings, to show the general usefulness of our proposed methods.

\begin{figure}[h]
    \vspace{-10pt}
    \begin{center}
        \begin{tabular}{cc}
          \rotatebox[origin=c]{90}{
            \begin{tabular}{cc}
                (b)  \quad \quad \ &
                (a) \quad  \
            \end{tabular}
          } &
         \hspace{-0.5cm}
         \raisebox{-.5\height}{\includegraphics[scale=0.50]{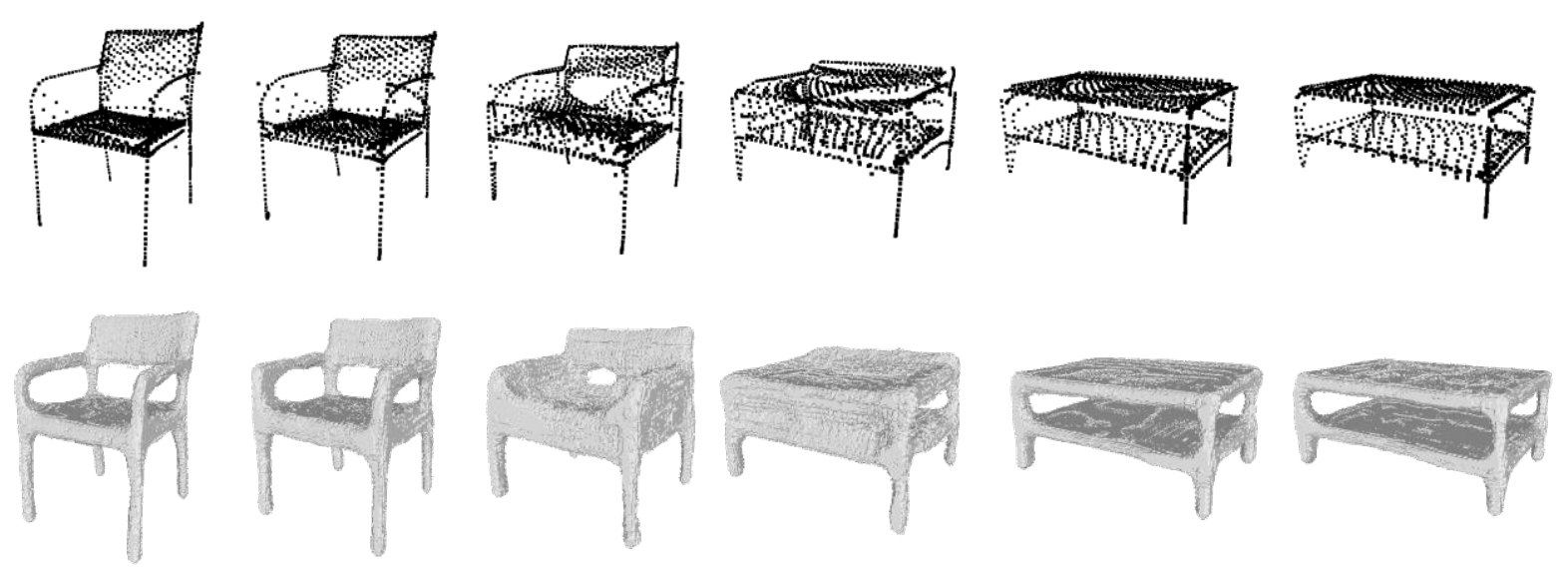}}
        \end{tabular}
        \caption{Example results of skeleton interpolation in the forms of skeletal point set (a) and skeletal volume (b).}
        \label{fig:interp}
    \end{center}
    \vspace{-10pt}
\end{figure}

\vspace{0.1cm}
\noindent \textbf{\emph{Interpolation of skeletons}}
SkeletonNet takes an RGB image of an object instance as input, and outputs its skeletons in both point set and volume based representations. In general, inputs of different instances correspond to different skeletal results. It is interesting to observe the results in between, i.e., the interpolations of skeletons between two results corresponding to two RGB inputs of different instances, possibly of different object categories. To this end, given two RGB inputs, we simply interpolate their global feature vectors from the image encoder, and then use the interpolated features for inference of skeletal point sets and the refined skeletal volumes. Example results in Fig.~\ref{fig:interp} show that, by using different weights of combination, the interpolated results demonstrate a gradual transition of plausible shapes from skeletons of \emph{chair} to those of \emph{table}.

\begin{figure}[t]
	\begin{center}
		\includegraphics[scale=0.30]{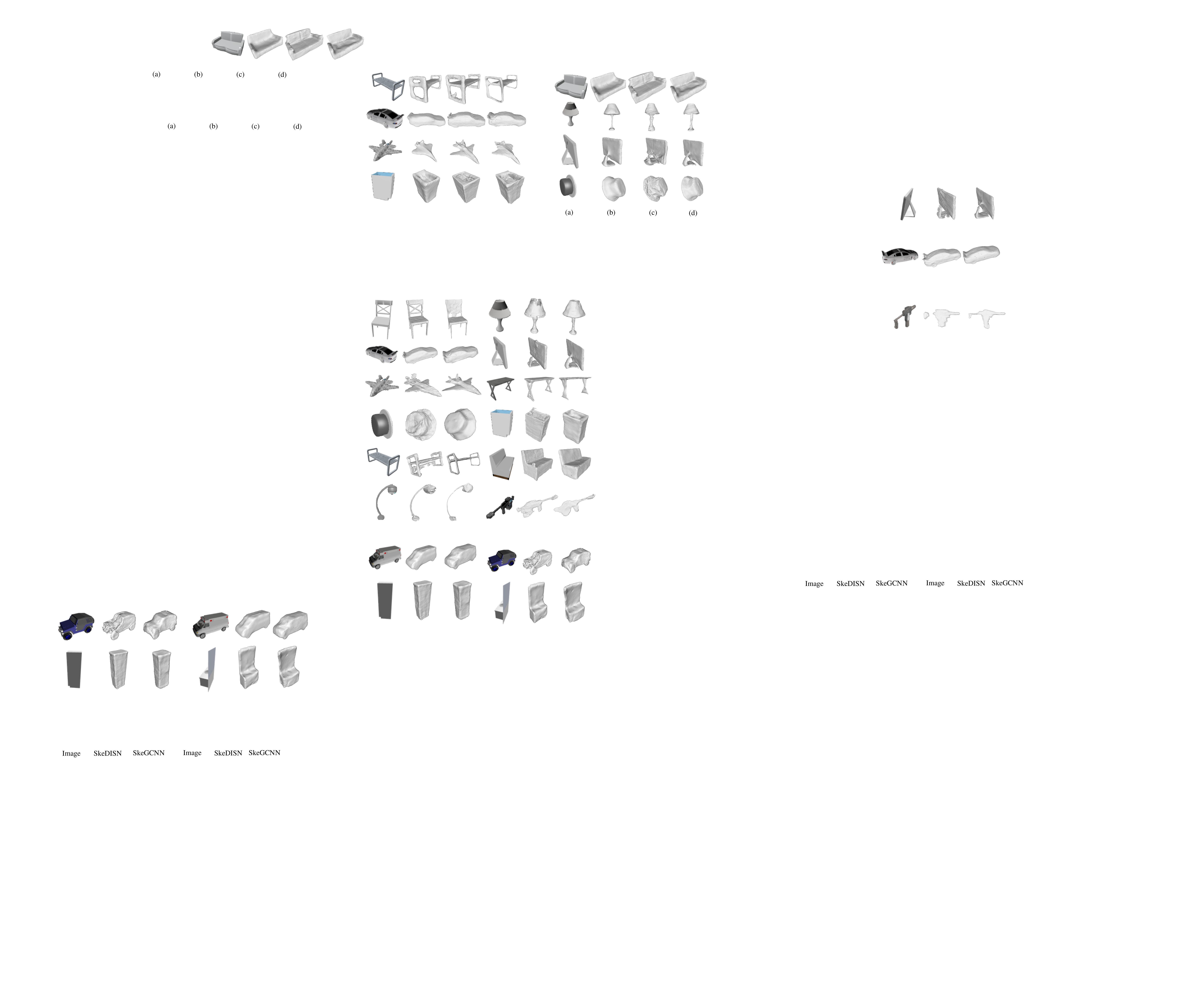}\\
		\begin{tabular}{p{30pt}p{30pt}p{30pt}p{30pt}p{30pt}p{30pt}}
                    \quad \ \ \ (a) &
                    \quad \ \ (b) &
                    \quad (c) &
                    \quad (a) &
                    \ \  (b) &
                     (c)
            \end{tabular}
        \vspace{-15pt}
		\caption{Additional results of explicit and implicit mesh recoveries given by our proposed SkeGCNN and SkeDISN. (a) Input images; (b) SkeDISN; (c) SkeGCNN. } 
		\label{fig:gallery}
	\end{center}
	\vspace{-15pt}
\end{figure}

\vspace{0.1cm}
\noindent \textbf{\emph{Result gallery}}
In Fig. \ref{fig:gallery}, we show additional results of our proposed SkeGCNN and SkeDISN respectively for explicit and implicit mesh recoveries. Generally speaking, the explicit method of SkeGCNN gives smoother mesh results, and the implicit method of SkeDISN gives results with more surface details, which may translate as improved quality for surface of complex topologies.

\begin{figure}[h]
    \begin{center}
    \includegraphics[scale=0.4]{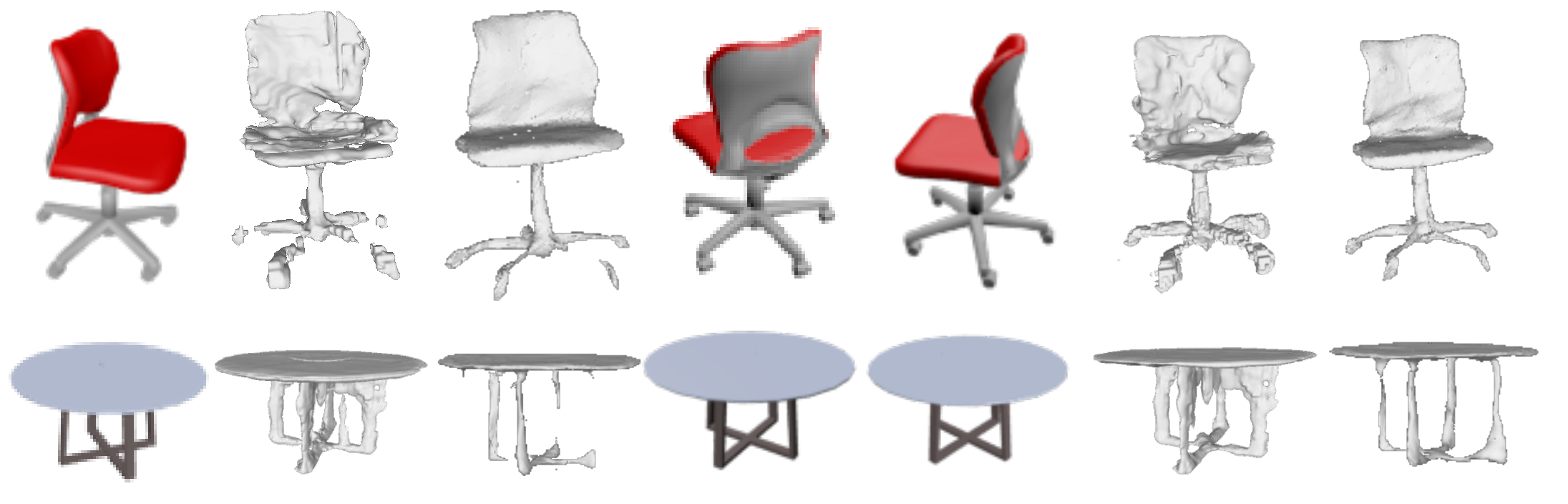}\\
    \begin{tabular}{p{25pt}p{25pt}p{25pt}p{25pt}p{25pt}p{25pt}p{25pt}}
                    \quad  (a)  &
                    \   (b)  &
                      (c)  &
                      (d) &
                      (e) &
                      (f) &
                      (g)
    \end{tabular}
    \caption{Example results of multi-view mesh recovery from our proposed SkeGCNN and SkeDISN.(a) View1; (b) SkeDISN; (c) SkeGCNN; (d) View2; (e) View3; (f) multi-view version of SkeDISN; (g)  multi-view version of SkeGCNN; }
    \label{fig:multi_view}	
    \end{center}
    \vspace{-15pt}
\end{figure}

\begin{figure}[h]
    \begin{center}
        \begin{tabular}{cc}
          \rotatebox[origin=c]{90}{
             \begin{tabular}{ccc}
                (c)  \quad  \quad \quad &
                (b)  \quad  \quad  \quad  &
                (a)  \quad
            \end{tabular}
          } &
         \hspace{-0.5cm}
         \raisebox{-.5\height}{\includegraphics[scale=0.35]{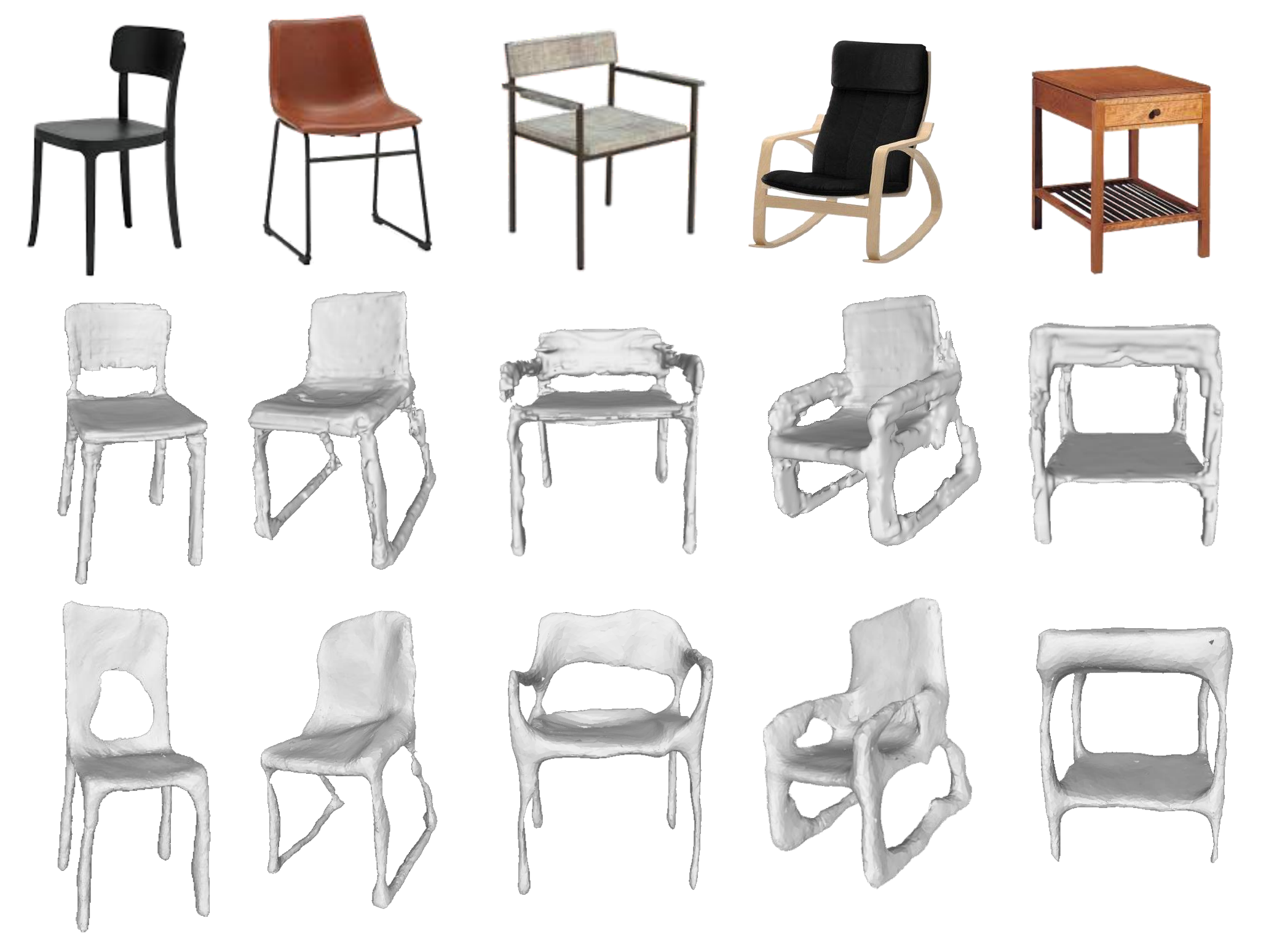}}
        \end{tabular}
        \caption{Example results of SkeGCNN and SkeDISN on novel instances contained in real images of Stanford online product images. (a) Input images; (b) SkeDISN; (c)SkeGCNN.}
        \label{fig:real}	
    \end{center}
    \vspace{-10pt}
\end{figure}

\vspace{0.1cm}
\noindent \textbf{\emph{Mesh Recovery from multi-view inputs}}
Our methods can be easily extended to take multiple RGB images of different views as inputs, and the results are expected to be improved. To achieve skeleton based mesh recovery from multi-view images, we modify SkeletonNet, SkeGCNN, and SkeDISN simply by conducting parallel feature encodings of the input views using their respective image encoders, and then aggregating the obtained global (and local) feature vectors via max pooling. Example results in Fig.~\ref{fig:multi_view} tell that for objects of complex topologies, using additional views help mesh recoveries of both explicit and implicit methods.

\vspace{0.1cm}
\noindent \textbf{\emph{Generalization on real images}}
We also validate our SkeDISN and SkeGCNN on Stanford online product images without further fine-tuning. Example results in Fig.~\ref{fig:real}  show that both our explicit and implicit methods are able to generalize to novel instances in real images.

\begin{figure}[h]
	\begin{center}
		\includegraphics[scale=0.54]{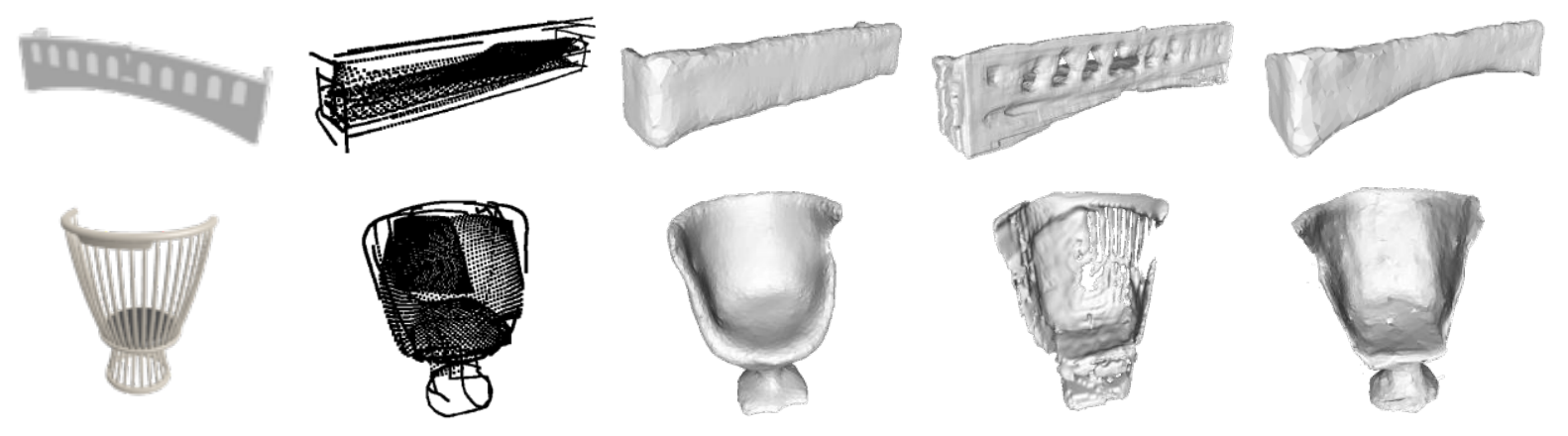}\\
		\begin{tabular}{p{35pt}p{35pt}p{35pt}p{35pt}p{35pt}}
                     \ \ (a) &
                     \quad (b)  &
                     \quad (c)  &
                     \quad \ \ (d) &
                     \quad \ \ (e)
        \end{tabular}
		\caption{Failure cases. (a) Input images; (b) The produced skeletal points; (c) The refined skeletal volumes; (d) SkeDISN; (e) SkeGCNN.}
		\label{fig:failure}
	\end{center}
	\vspace{-20pt}
\end{figure}

\noindent \textbf{\emph{Failure Cases}}
Although SkeletonNet can preserve long and thin structures, it still has difficulty in capturing topologies of those objects with tiny holes, such as the chair and bench shown in Fig.~\ref{fig:failure}.

\begin{figure}[h]
	\begin{center}
		\includegraphics[scale=0.375]{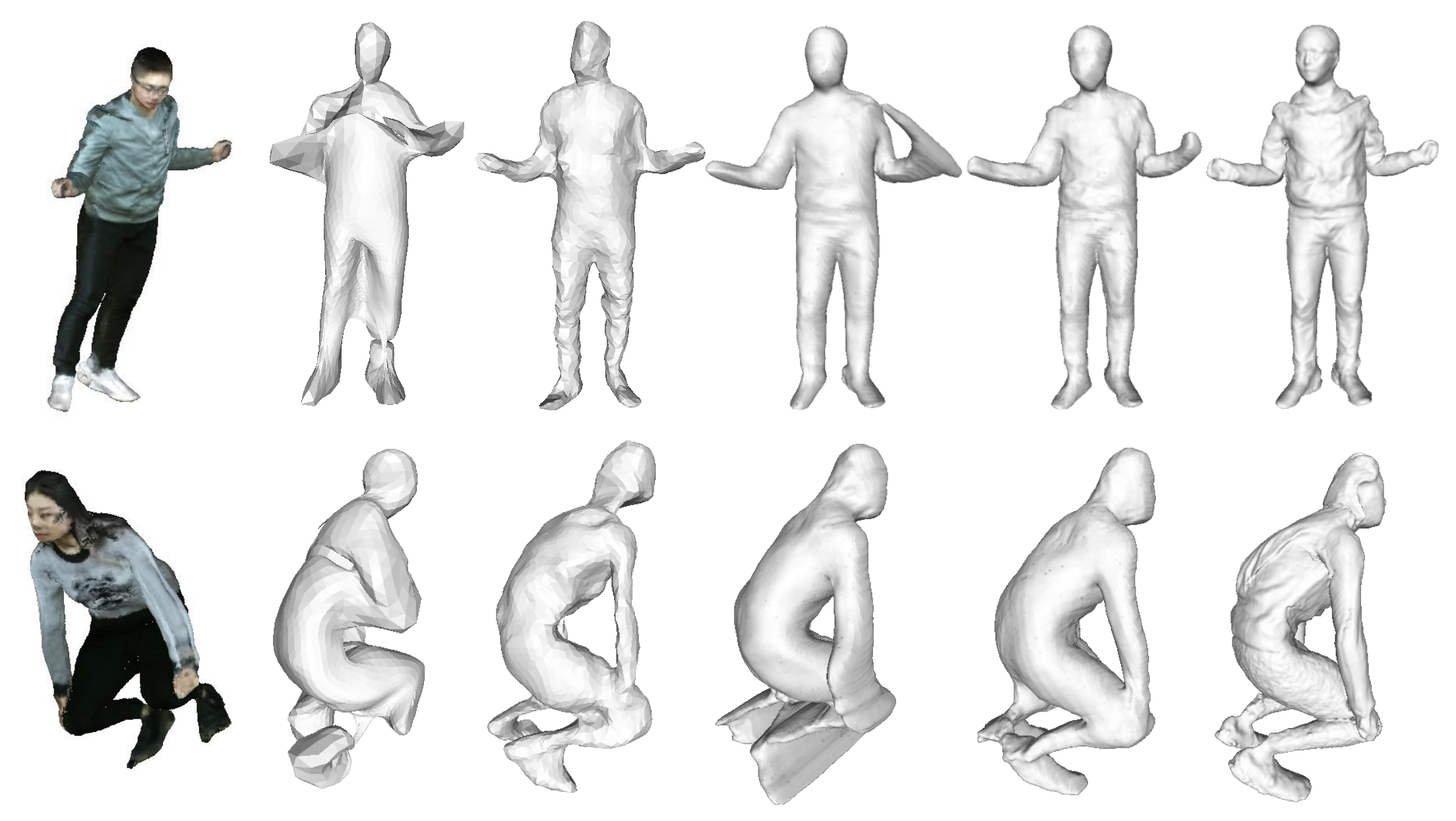}\\
		 \begin{tabular}{p{30pt}p{30pt}p{30pt}p{30pt}p{30pt}p{30pt}}
                \quad \ \ (a) &
                \quad  (b) &
                \   (c) &
                \   (d) &
                \   (e) &
                  (f)
        \end{tabular}
		\caption{Qualitative results on the tasks of explicit and implicit human mesh recoveries. (a) Input images; (b) Pixel2Mesh; (c) SkeGCNN;  (d) PIFu; (e) SkePIFu; (f) Ground Truths.}
		\label{fig:deephuman}
	\end{center}
	\vspace{-20pt}
\end{figure}

\vspace{0.1cm}
\noindent \textbf{\emph{Human Body Recovery}}
We focus on surface recoveries of man-made objects in the present paper. We note that our method can also utilize the skeleton to represent articulated human bodies. To do so, we conduct an extra experiment and apply our method to single-view 3D human reconstruction, where we use the DeepHuman dataset~\cite{zheng2019deephuman}. Example results are shown in Fig.~\ref{fig:skehuman}. The results demonstrate that our SkeletonNet can capture the structures of human shapes with different poses. We also compare with existing methods to validate the effectiveness of SkeletonNet for the tasks of both explicit and implicit mesh recoveries of human bodies. For explicit mesh recovery, we compare our SkeGCNN with Pixel2Mesh, which deforms the SMPL template mesh~\cite{loper2015smpl} under the rest pose to fit the target human surface mesh using a graph CNN.  For implicit mesh recovery, we compare with the representative method of PIFu~\cite{saito2019pifu}, which predicts pixel-aligned implicit functions based on projected 2D image features; our newly proposed SkePIFu primarily follows PIFu and is based on Skeleton-Regularized Pixel-aligned Implicit Function learning. More specifically, SkePIFu regularizes the learning of a binary occupancy field using multi-scale, local features of the skeletal volume.
As seen in Fig.~\ref{fig:deephuman}, SkeletonNet indeed improves the explicit and implicit human mesh recoveries, which is also verified by the quantitative comparisons in Table.~\ref{tab:deephuman}.

\begin{table}[h]
    \vspace{-5pt}
	\begin{center}
		\begin{tabular}{c | c  c}\hline
			\centering
			Method & CD ($\times 0.0001$) & IoU ($\times 100$) \\
			\hline
			Pixel2Mesh &  8.358  &  50.77 \\
			SkeGCNN   & \textbf{4.294}  & \textbf{68.37} \\
		    \hline
			PIFu  & 2.990  &  65.86\\
			SkePIFu & \textbf{2.329}  & \textbf{69.02} \\
			\hline
		\end{tabular}
		\caption{Quantitative results on the tasks of explicit and implicit human mesh recoveries. Results are obtained on the DeepHuman~\cite{zheng2019deephuman} dataset.}
	    \label{tab:deephuman}
	\end{center}
	\vspace{-15pt}
\end{table}

\section{Conclusion}

    Recovering the surface shape of an object from one or a few perspectives of images is a fundamental yet challenging computer vision task. This work proposes to learn the topology-preserved, skeletal shape representation, and uses the learned representation to assist the surface recovery task of interest. In order to learn the skeletal representation, we propose an end-to-end, trainable \emph{SkeletonNet} that is able to generate a high-quality skeletal volume via a bridged learning of a skeletal point set. The obtained skeletal volume can be used either as a bridge to explicitly recover a surface mesh in our proposed SkeGCNN, or as a constraint to regularize the learning of an implicit surface field in our proposed SkeDISN. Thorough experiments confirm both the efficacy of SkeletonNet for producing high-quality skeletal representations, and its usefulness for the downstream surface recovery tasks.

    However, similar to most of existing mesh reconstruction methods, our proposed approaches are limited in dealing with diverse natural images in the wild, since our models are trained on synthetic image datasets. Due to the varying materials, textures, and/or lighting conditions, there exists a domain gap between synthetic and natural images. In fact, research topics such as photorealistic rendering~\cite{su2015render, yariv2020multiview} or domain adaptation~\cite{pinheiro2019domain} are particularly for addressing such issues.
    In future research, we will pursue surface reconstruction from more natural images. We are also interested in discovering the usefulness of skeletal representations for other shape analysis and reconstruction tasks, such as shape classification, segmentation, and completion.

\ifCLASSOPTIONcompsoc
  \section*{Acknowledgments}
\else
  \section*{Acknowledgment}
\fi

 This work is supported in part by the National Natural Science Foundation of China (Grant No.: 61771201, 61629101, 61902334), the Program for Guangdong Introducing Innovative and Enterpreneurial Teams (Grant No.: 2017ZT07X183), the Guangdong R$\&$D key project of China (Grant No.: 2019B010155001), the Key Area R$\&$D Program of Guangdong Province (Grant No.: 2018B030338001), the National Key R$\&$D Program of China (Grant No.: 2018YFB1800800), Guangdong Research Project (Grant No.: 2017ZT07X152), Shenzhen Key Lab Fund (Grant No.: ZDSYS201707251409055), and Microsoft Research Asia.

\bibliographystyle{IEEEtran}
\bibliography{refer}

\begin{thebibliography}{10}
\providecommand{\url}[1]{#1}
\csname url@samestyle\endcsname
\providecommand{\newblock}{\relax}
\providecommand{\bibinfo}[2]{#2}
\providecommand{\BIBentrySTDinterwordspacing}{\spaceskip=0pt\relax}
\providecommand{\BIBentryALTinterwordstretchfactor}{4}
\providecommand{\BIBentryALTinterwordspacing}{\spaceskip=\fontdimen2\font plus
\BIBentryALTinterwordstretchfactor\fontdimen3\font minus
  \fontdimen4\font\relax}
\providecommand{\BIBforeignlanguage}[2]{{%
\expandafter\ifx\csname l@#1\endcsname\relax
\typeout{** WARNING: IEEEtran.bst: No hyphenation pattern has been}%
\typeout{** loaded for the language `#1'. Using the pattern for}%
\typeout{** the default language instead.}%
\else
\language=\csname l@#1\endcsname
\fi
#2}}
\providecommand{\BIBdecl}{\relax}
\BIBdecl

\bibitem{marr1979computational}
D.~Marr and T.~Poggio, ``A computational theory of human stereo vision,''
  \emph{Proceedings of the Royal Society of London. Series B. Biological
  Sciences}, vol. 204, no. 1156, pp. 301--328, 1979.

\bibitem{hartley2000multiple}
R.~{Hartley} and A.~{Zisserman}, \emph{Multiple View Geometry in Computer
  Vision}, 2000.

\bibitem{hirschmuller2007stereo}
H.~Hirschmuller, ``Stereo processing by semiglobal matching and mutual
  information,'' \emph{IEEE Transactions on pattern analysis and machine
  intelligence}, vol.~30, no.~2, pp. 328--341, 2007.

\bibitem{campbell2008using}
N.~D. Campbell, G.~Vogiatzis, C.~Hern{\'a}ndez, and R.~Cipolla, ``Using
  multiple hypotheses to improve depth-maps for multi-view stereo,'' in
  \emph{European Conference on Computer Vision}.\hskip 1em plus 0.5em minus
  0.4em\relax Springer, 2008, pp. 766--779.

\bibitem{tola2012efficient}
E.~Tola, C.~Strecha, and P.~Fua, ``Efficient large-scale multi-view stereo for
  ultra high-resolution image sets,'' \emph{Machine Vision and Applications},
  vol.~23, no.~5, pp. 903--920, 2012.

\bibitem{merrell2007real}
P.~Merrell, A.~Akbarzadeh, L.~Wang, P.~Mordohai, J.-M. Frahm, R.~Yang,
  D.~Nist{\'e}r, and M.~Pollefeys, ``Real-time visibility-based fusion of depth
  maps,'' in \emph{2007 IEEE 11th International Conference on Computer
  Vision}.\hskip 1em plus 0.5em minus 0.4em\relax IEEE, 2007, pp. 1--8.

\bibitem{zach2007globally}
C.~Zach, T.~Pock, and H.~Bischof, ``A globally optimal algorithm for robust
  tv-l 1 range image integration,'' in \emph{2007 IEEE 11th International
  Conference on Computer Vision}.\hskip 1em plus 0.5em minus 0.4em\relax IEEE,
  2007, pp. 1--8.

\bibitem{newcombe2011kinectfusion}
R.~A. Newcombe, S.~Izadi, O.~Hilliges, D.~Molyneaux, D.~Kim, A.~J. Davison,
  P.~Kohi, J.~Shotton, S.~Hodges, and A.~Fitzgibbon, ``Kinectfusion: Real-time
  dense surface mapping and tracking,'' in \emph{2011 10th IEEE International
  Symposium on Mixed and Augmented Reality}.\hskip 1em plus 0.5em minus
  0.4em\relax IEEE, 2011, pp. 127--136.

\bibitem{choy20163d}
C.~B. {Choy}, D.~{Xu}, J.~{Gwak}, K.~{Chen}, and S.~{Savarese}, ``3d-r2n2: A
  unified approach for single and multi-view 3d object reconstruction,''
  \emph{european conference on computer vision}, pp. 628--644, 2016.

\bibitem{girdhar2016learning}
R.~Girdhar, D.~F. Fouhey, M.~Rodriguez, and A.~Gupta, ``Learning a predictable
  and generative vector representation for objects,'' in \emph{European
  Conference on Computer Vision}.\hskip 1em plus 0.5em minus 0.4em\relax
  Springer, 2016, pp. 484--499.

\bibitem{tatarchenko2017octree}
M.~{Tatarchenko}, A.~{Dosovitskiy}, and T.~{Brox}, ``Octree generating
  networks: Efficient convolutional architectures for high-resolution 3d
  outputs,'' in \emph{2017 IEEE International Conference on Computer Vision
  (ICCV)}, 2017, pp. 2107--2115.

\bibitem{groueix2018atlasnet}
T.~{Groueix}, M.~{Fisher}, V.~G. {Kim}, B.~C. {Russell}, and M.~{Aubry},
  ``Atlasnet: A papier-mâché approach to learning 3d surface generation,''
  \emph{computer vision and pattern recognition}, 2018.

\bibitem{kato2018neural}
H.~{Kato}, Y.~{Ushiku}, and T.~{Harada}, ``Neural 3d mesh renderer,''
  \emph{computer vision and pattern recognition}, pp. 3907--3916, 2018.

\bibitem{fan2017point}
H.~{Fan}, H.~{Su}, and L.~J. {Guibas}, ``A point set generation network for 3d
  object reconstruction from a single image,'' in \emph{2017 IEEE Conference on
  Computer Vision and Pattern Recognition (CVPR)}, 2017, pp. 2463--2471.

\bibitem{wang2018pixel2mesh}
N.~{Wang}, Y.~{Zhang}, Z.~{Li}, Y.~{Fu}, W.~{Liu}, and Y.-G. {Jiang},
  ``Pixel2mesh: Generating 3d mesh models from single rgb images.'' \emph{arXiv
  preprint arXiv:1804.01654}, 2018.

\bibitem{tang2019skeleton}
J.~Tang, X.~Han, J.~Pan, K.~Jia, and X.~Tong, ``A skeleton-bridged deep
  learning approach for generating meshes of complex topologies from single rgb
  images,'' in \emph{Proceedings of the IEEE Conference on Computer Vision and
  Pattern Recognition}, 2019, pp. 4541--4550.

\bibitem{pan2018residual}
J.~Pan, J.~Li, X.~Han, and K.~Jia, ``Residual meshnet: Learning to deform
  meshes for single-view 3d reconstruction,'' in \emph{2018 International
  Conference on 3D Vision (3DV)}.\hskip 1em plus 0.5em minus 0.4em\relax IEEE,
  2018, pp. 719--727.

\bibitem{pan2019deep}
J.~Pan, X.~Han, W.~Chen, J.~Tang, and K.~Jia, ``Deep mesh reconstruction from
  single rgb images via topology modification networks,'' in \emph{Proceedings
  of the IEEE International Conference on Computer Vision}, 2019, pp.
  9964--9973.

\bibitem{michalkiewicz2019deep}
M.~Michalkiewicz, J.~K. Pontes, D.~Jack, M.~Baktashmotlagh, and A.~Eriksson,
  ``Deep level sets: Implicit surface representations for 3d shape inference,''
  \emph{arXiv preprint arXiv:1901.06802}, 2019.

\bibitem{mescheder2019occupancy}
L.~Mescheder, M.~Oechsle, M.~Niemeyer, S.~Nowozin, and A.~Geiger, ``Occupancy
  networks: Learning 3d reconstruction in function space,'' in
  \emph{Proceedings of the IEEE Conference on Computer Vision and Pattern
  Recognition}, 2019, pp. 4460--4470.

\bibitem{chen2019learning}
Z.~Chen and H.~Zhang, ``Learning implicit fields for generative shape
  modeling,'' in \emph{Proceedings of the IEEE Conference on Computer Vision
  and Pattern Recognition}, 2019, pp. 5939--5948.

\bibitem{park2019deepsdf}
J.~J. Park, P.~Florence, J.~Straub, R.~Newcombe, and S.~Lovegrove, ``Deepsdf:
  Learning continuous signed distance functions for shape representation,'' in
  \emph{Proceedings of the IEEE Conference on Computer Vision and Pattern
  Recognition}, 2019, pp. 165--174.

\bibitem{xu2019disn}
Q.~Xu, W.~Wang, D.~Ceylan, R.~Mech, and U.~Neumann, ``Disn: Deep implicit
  surface network for high-quality single-view 3d reconstruction,'' \emph{arXiv
  preprint arXiv:1905.10711}, 2019.

\bibitem{wu2016learning}
J.~Wu, C.~Zhang, T.~Xue, B.~Freeman, and J.~Tenenbaum, ``Learning a
  probabilistic latent space of object shapes via 3d generative-adversarial
  modeling,'' in \emph{Advances in Neural Information Processing Systems},
  2016, pp. 82--90.

\bibitem{simonyan2015very}
K.~{Simonyan} and A.~{Zisserman}, ``Very deep convolutional networks for
  large-scale image recognition,'' \emph{international conference on learning
  representations}, 2015.

\bibitem{krizhevsky2012imagenet}
A.~{Krizhevsky}, I.~{Sutskever}, and G.~E. {Hinton}, ``Imagenet classification
  with deep convolutional neural networks,'' in \emph{Advances in Neural
  Information Processing Systems 25}, 2012, pp. 1097--1105.

\bibitem{szegedy2015going}
C.~{Szegedy}, W.~{Liu}, Y.~{Jia}, P.~{Sermanet}, S.~E. {Reed}, D.~{Anguelov},
  D.~{Erhan}, V.~{Vanhoucke}, and A.~{Rabinovich}, ``Going deeper with
  convolutions,'' in \emph{2015 IEEE Conference on Computer Vision and Pattern
  Recognition (CVPR)}, 2015, pp. 1--9.

\bibitem{he2016deep}
K.~He, X.~Zhang, S.~Ren, and J.~Sun, ``Deep residual learning for image
  recognition,'' in \emph{Proceedings of the IEEE conference on computer vision
  and pattern recognition}, 2016, pp. 770--778.

\bibitem{riegler2017octnet}
G.~Riegler, A.~O. Ulusoy, and A.~Geiger, ``Octnet: Learning deep 3d
  representations at high resolutions,'' in \emph{Proceedings of the IEEE
  Conference on Computer Vision and Pattern Recognition}, vol.~3, 2017.

\bibitem{riegler2017octnetfusion}
G.~Riegler, A.~O. Ulusoy, H.~Bischof, and A.~Geiger, ``Octnetfusion: Learning
  depth fusion from data,'' in \emph{2017 International Conference on 3D Vision
  (3DV)}.\hskip 1em plus 0.5em minus 0.4em\relax IEEE, 2017, pp. 57--66.

\bibitem{hane2017hierarchical}
C.~H{\"a}ne, S.~Tulsiani, and J.~Malik, ``Hierarchical surface prediction for
  3d object reconstruction,'' in \emph{2017 International Conference on 3D
  Vision (3DV)}.\hskip 1em plus 0.5em minus 0.4em\relax IEEE, 2017, pp.
  412--420.

\bibitem{wang2017cnn}
P.-S. Wang, Y.~Liu, Y.-X. Guo, C.-Y. Sun, and X.~Tong, ``O-cnn: Octree-based
  convolutional neural networks for 3d shape analysis,'' \emph{ACM Transactions
  on Graphics (TOG)}, vol.~36, no.~4, p.~72, 2017.

\bibitem{wang2018adaptive}
P.-S. Wang, C.-Y. Sun, Y.~Liu, and X.~Tong, ``Adaptive o-cnn: a patch-based
  deep representation of 3d shapes,'' in \emph{SIGGRAPH Asia 2018 Technical
  Papers}.\hskip 1em plus 0.5em minus 0.4em\relax ACM, 2018, p. 217.

\bibitem{lorensen1987marching}
W.~{Lorensen} and H.~E. {Cline}, ``Marching cubes: A high resolution 3d surface
  construction algorithm,'' \emph{Computers Graphics}, 1987.

\bibitem{arsalan2017synthesizing}
A.~Arsalan~Soltani, H.~Huang, J.~Wu, T.~D. Kulkarni, and J.~B. Tenenbaum,
  ``Synthesizing 3d shapes via modeling multi-view depth maps and silhouettes
  with deep generative networks,'' in \emph{Proceedings of the IEEE conference
  on computer vision and pattern recognition}, 2017, pp. 1511--1519.

\bibitem{lin2018learning}
C.-H. Lin, C.~Kong, and S.~Lucey, ``Learning efficient point cloud generation
  for dense 3d object reconstruction,'' in \emph{Thirty-Second AAAI Conference
  on Artificial Intelligence}, 2018.

\bibitem{tatarchenko2016multi}
M.~Tatarchenko, A.~Dosovitskiy, and T.~Brox, ``Multi-view 3d models from single
  images with a convolutional network,'' in \emph{European Conference on
  Computer Vision}.\hskip 1em plus 0.5em minus 0.4em\relax Springer, 2016, pp.
  322--337.

\bibitem{tagliasacchi2012mean}
A.~Tagliasacchi, I.~Alhashim, M.~Olson, and H.~Zhang, ``Mean curvature
  skeletons,'' in \emph{Computer Graphics Forum}, vol.~31, no.~5.\hskip 1em
  plus 0.5em minus 0.4em\relax Wiley Online Library, 2012, pp. 1735--1744.

\bibitem{wu2015deep}
S.~Wu, H.~Huang, M.~Gong, M.~Zwicker, and D.~Cohen-Or, ``Deep points
  consolidation,'' \emph{ACM Transactions on Graphics (TOG)}, vol.~34, no.~6,
  p. 176, 2015.

\bibitem{chang2015shapenet}
A.~X. Chang, T.~Funkhouser, L.~Guibas, P.~Hanrahan, Q.~Huang, Z.~Li,
  S.~Savarese, M.~Savva, S.~Song, H.~Su \emph{et~al.}, ``Shapenet: An
  information-rich 3d model repository,'' \emph{arXiv preprint
  arXiv:1512.03012}, 2015.

\bibitem{jiang2018gal}
L.~Jiang, S.~Shi, X.~Qi, and J.~Jia, ``Gal: Geometric adversarial loss for
  single-view 3d-object reconstruction,'' in \emph{Proceedings of the European
  Conference on Computer Vision (ECCV)}, 2018, pp. 802--816.

\bibitem{achlioptas2018learning}
P.~{Achlioptas}, O.~{Diamanti}, I.~{Mitliagkas}, and L.~{Guibas}, ``Learning
  representations and generative models for 3d point clouds,'' in \emph{ICLR
  2018 : International Conference on Learning Representations 2018}, 2018.

\bibitem{Williams_2019_CVPR}
F.~Williams, T.~Schneider, C.~Silva, D.~Zorin, J.~Bruna, and D.~Panozzo, ``Deep
  geometric prior for surface reconstruction,'' in \emph{Proceedings of the
  IEEE/CVF Conference on Computer Vision and Pattern Recognition (CVPR)}, June
  2019.

\bibitem{jack2018learning}
D.~Jack, J.~K. Pontes, S.~Sridharan, C.~Fookes, S.~Shirazi, F.~Maire, and
  A.~Eriksson, ``Learning free-form deformations for 3d object
  reconstruction,'' in \emph{Asian Conference on Computer Vision}.\hskip 1em
  plus 0.5em minus 0.4em\relax Springer, 2018, pp. 317--333.

\bibitem{pontes2018image2mesh}
J.~K. Pontes, C.~Kong, S.~Sridharan, S.~Lucey, A.~Eriksson, and C.~Fookes,
  ``Image2mesh: A learning framework for single image 3d reconstruction,'' in
  \emph{Asian Conference on Computer Vision}.\hskip 1em plus 0.5em minus
  0.4em\relax Springer, 2018, pp. 365--381.

\bibitem{wang20193dn}
W.~Wang, D.~Ceylan, R.~Mech, and U.~Neumann, ``3dn: 3d deformation network,''
  in \emph{Proceedings of the IEEE Conference on Computer Vision and Pattern
  Recognition}, 2019, pp. 1038--1046.

\bibitem{nie2020total3dunderstanding}
Y.~{Nie}, X.~{Han}, S.~{Guo}, Y.~{Zheng}, J.~{Chang}, and J.~J. {Zhang},
  ``Total3dunderstanding: Joint layout, object pose and mesh reconstruction for
  indoor scenes from a single image.'' \emph{arXiv preprint arXiv:2002.12212},
  2020.

\bibitem{hanocka2020point2mesh}
R.~Hanocka, G.~Metzer, R.~Giryes, and D.~Cohen-Or, ``Point2mesh: a self-prior
  for deformable meshes,'' \emph{arXiv preprint arXiv:2005.11084}, 2020.

\bibitem{kar2017learning}
A.~Kar, C.~H{\"a}ne, and J.~Malik, ``Learning a multi-view stereo machine,'' in
  \emph{Advances in neural information processing systems}, 2017, pp. 365--376.

\bibitem{goodfellow2014generative}
I.~Goodfellow, J.~Pouget-Abadie, M.~Mirza, B.~Xu, D.~Warde-Farley, S.~Ozair,
  A.~Courville, and Y.~Bengio, ``Generative adversarial nets,'' in
  \emph{Advances in neural information processing systems}, 2014, pp.
  2672--2680.

\bibitem{chen2020bsp}
Z.~Chen, A.~Tagliasacchi, and H.~Zhang, ``Bsp-net: Generating compact meshes
  via binary space partitioning,'' in \emph{Proceedings of the IEEE/CVF
  Conference on Computer Vision and Pattern Recognition}, 2020, pp. 45--54.

\bibitem{lei2020analytic}
J.~Lei and K.~Jia, ``Analytic marching: An analytic meshing solution from deep
  implicit surface networks,'' \emph{arXiv preprint arXiv:2002.06597}, 2020.

\bibitem{atzmon2020sal}
M.~Atzmon and Y.~Lipman, ``Sal: Sign agnostic learning of shapes from raw
  data,'' in \emph{Proceedings of the IEEE/CVF Conference on Computer Vision
  and Pattern Recognition}, 2020, pp. 2565--2574.

\bibitem{gropp2020implicit}
A.~Gropp, L.~Yariv, N.~Haim, M.~Atzmon, and Y.~Lipman, ``Implicit geometric
  regularization for learning shapes,'' \emph{arXiv preprint arXiv:2002.10099},
  2020.

\bibitem{chibane2020implicit}
J.~Chibane, T.~Alldieck, and G.~Pons-Moll, ``Implicit functions in feature
  space for 3d shape reconstruction and completion,'' in \emph{Proceedings of
  the IEEE/CVF Conference on Computer Vision and Pattern Recognition}, 2020,
  pp. 6970--6981.

\bibitem{peng2020convolutional}
S.~Peng, M.~Niemeyer, L.~Mescheder, M.~Pollefeys, and A.~Geiger,
  ``Convolutional occupancy networks,'' \emph{arXiv preprint arXiv:2003.04618},
  vol.~2, 2020.

\bibitem{saito2019pifu}
S.~Saito, Z.~Huang, R.~Natsume, S.~Morishima, A.~Kanazawa, and H.~Li, ``Pifu:
  Pixel-aligned implicit function for high-resolution clothed human
  digitization,'' \emph{arXiv preprint arXiv:1905.05172}, 2019.

\bibitem{gao2020learning}
J.~Gao, W.~Chen, T.~Xiang, A.~Jacobson, M.~McGuire, and S.~Fidler, ``Learning
  deformable tetrahedral meshes for 3d reconstruction,'' \emph{arXiv preprint
  arXiv:2011.01437}, 2020.

\bibitem{oechsle2019texture}
M.~Oechsle, L.~Mescheder, M.~Niemeyer, T.~Strauss, and A.~Geiger, ``Texture
  fields: Learning texture representations in function space,'' in
  \emph{Proceedings of the IEEE/CVF International Conference on Computer
  Vision}, 2019, pp. 4531--4540.

\bibitem{oechsle2020learning}
M.~Oechsle, M.~Niemeyer, C.~Reiser, L.~Mescheder, T.~Strauss, and A.~Geiger,
  ``Learning implicit surface light fields,'' in \emph{2020 International
  Conference on 3D Vision (3DV)}.\hskip 1em plus 0.5em minus 0.4em\relax IEEE,
  2020, pp. 452--462.

\bibitem{yariv2020multiview}
L.~Yariv, Y.~Kasten, D.~Moran, M.~Galun, M.~Atzmon, B.~Ronen, and Y.~Lipman,
  ``Multiview neural surface reconstruction by disentangling geometry and
  appearance,'' \emph{Advances in Neural Information Processing Systems},
  vol.~33, 2020.

\bibitem{blum1967transformation}
H.~Blum, \emph{A transformation for extracting new descriptors of shape}.\hskip
  1em plus 0.5em minus 0.4em\relax MIT press Cambridge, 1967, vol.~4.

\bibitem{siddiqi2008medial}
K.~Siddiqi and S.~Pizer, \emph{Medial representations: mathematics, algorithms
  and applications}.\hskip 1em plus 0.5em minus 0.4em\relax Springer Science \&
  Business Media, 2008, vol.~37.

\bibitem{attali1996modeling}
D.~Attali and A.~Montanvert, ``Modeling noise for a better simplification of
  skeletons,'' in \emph{Proceedings of 3rd IEEE International Conference on
  Image Processing}, vol.~3.\hskip 1em plus 0.5em minus 0.4em\relax IEEE, 1996,
  pp. 13--16.

\bibitem{chazal2005lambda}
F.~Chazal and A.~Lieutier, ``The “$\lambda$-medial axis”,'' \emph{Graphical
  Models}, vol.~67, no.~4, pp. 304--331, 2005.

\bibitem{miklos2010discrete}
B.~Miklos, J.~Giesen, and M.~Pauly, ``Discrete scale axis representations for
  3d geometry,'' in \emph{ACM SIGGRAPH 2010 papers}, 2010, pp. 1--10.

\bibitem{li2015q}
P.~Li, B.~Wang, F.~Sun, X.~Guo, C.~Zhang, and W.~Wang, ``Q-mat: Computing
  medial axis transform by quadratic error minimization,'' \emph{ACM
  Transactions on Graphics (TOG)}, vol.~35, no.~1, pp. 1--16, 2015.

\bibitem{pan2019q}
Y.~Pan, B.~Wang, X.~Guo, H.~Zeng, Y.~Ma, and W.~Wang, ``Q-mat+: An
  error-controllable and feature-sensitive simplification algorithm for medial
  axis transform,'' \emph{Computer Aided Geometric Design}, vol.~71, pp.
  16--29, 2019.

\bibitem{yang2020p2mat}
B.~Yang, J.~Yao, B.~Wang, J.~Hu, Y.~Pan, T.~Pan, W.~Wang, and X.~Guo,
  ``P2mat-net: Learning medial axis transform from sparse point clouds,''
  \emph{Computer Aided Geometric Design}, p. 101874, 2020.

\bibitem{hu2019mat}
J.~Hu, B.~Wang, L.~Qian, Y.~Pan, X.~Guo, L.~Liu, and W.~Wang, ``Mat-net: Medial
  axis transform network for 3d object recognition.'' in \emph{IJCAI}, 2019,
  pp. 774--781.

\bibitem{cornea2005curve}
N.~D. Cornea, D.~Silver, and P.~Min, ``Curve-skeleton applications,'' in
  \emph{VIS 05. IEEE Visualization, 2005.}\hskip 1em plus 0.5em minus
  0.4em\relax IEEE, 2005, pp. 95--102.

\bibitem{sharf2007fly}
A.~Sharf, T.~Lewiner, A.~Shamir, and L.~Kobbelt, ``On-the-fly curve-skeleton
  computation for 3d shapes,'' in \emph{Computer Graphics Forum}, vol.~26,
  no.~3.\hskip 1em plus 0.5em minus 0.4em\relax Wiley Online Library, 2007, pp.
  323--328.

\bibitem{tagliasacchi2009curve}
A.~Tagliasacchi, H.~Zhang, and D.~Cohen-Or, ``Curve skeleton extraction from
  incomplete point cloud,'' in \emph{ACM SIGGRAPH 2009 papers}, 2009, pp. 1--9.

\bibitem{bucksch2010skeltre}
A.~Bucksch, R.~Lindenbergh, and M.~Menenti, ``Skeltre,'' \emph{The Visual
  Computer}, vol.~26, no.~10, pp. 1283--1300, 2010.

\bibitem{huang2013l1}
H.~Huang, S.~Wu, D.~Cohen-Or, M.~Gong, H.~Zhang, G.~Li, and B.~Chen,
  ``L1-medial skeleton of point cloud.'' \emph{ACM Transactions on Graphics
  (TOG)}, vol.~32, no.~4, pp. 65--1, 2013.

\bibitem{yin2018p2p}
K.~Yin, H.~Huang, D.~Cohen-Or, and H.~Zhang, ``P2p-net: Bidirectional point
  displacement net for shape transform,'' \emph{ACM Transactions on Graphics
  (TOG)}, vol.~37, no.~4, pp. 1--13, 2018.

\bibitem{nie2020skeleton}
Y.~Nie, Y.~Lin, X.~Han, S.~Guo, J.~Chang, S.~Cui, and J.~J. Zhang,
  ``Skeleton-bridged point completion: From global inference to local
  adjustment,'' \emph{arXiv preprint arXiv:2010.07428}, 2020.

\bibitem{lin2020point2skeleton}
C.~Lin, C.~Li, Y.~Liu, N.~Chen, Y.-K. Choi, and W.~Wang, ``Point2skeleton:
  Learning skeletal representations from point clouds,'' \emph{arXiv preprint
  arXiv:2012.00230}, 2020.

\bibitem{wang2020pie}
X.~Wang, Y.~Xu, K.~Xu, A.~Tagliasacchi, B.~Zhou, A.~Mahdavi-Amiri, and
  H.~Zhang, ``Pie-net: Parametric inference of point cloud edges,'' \emph{arXiv
  preprint arXiv:2007.04883}, 2020.

\bibitem{field1988laplacian}
D.~A. Field, ``Laplacian smoothing and delaunay triangulations,''
  \emph{Communications in applied numerical methods}, vol.~4, no.~6, pp.
  709--712, 1988.

\bibitem{han2017high}
X.~{Han}, Z.~{Li}, H.~{Huang}, E.~{Kalogerakis}, and Y.~{Yu}, ``High-resolution
  shape completion using deep neural networks for global structure and local
  geometry inference,'' in \emph{2017 IEEE International Conference on Computer
  Vision (ICCV)}, 2017, pp. 85--93.

\bibitem{ronneberger2015u}
O.~{Ronneberger}, P.~{Fischer}, and T.~{Brox}, ``U-net: Convolutional networks
  for biomedical image segmentation,'' \emph{medical image computing and
  computer assisted intervention}, pp. 234--241, 2015.

\bibitem{zheng2019deephuman}
Z.~Zheng, T.~Yu, Y.~Wei, Q.~Dai, and Y.~Liu, ``Deephuman: 3d human
  reconstruction from a single image,'' in \emph{Proceedings of the IEEE/CVF
  International Conference on Computer Vision}, 2019, pp. 7739--7749.

\bibitem{loper2015smpl}
M.~Loper, N.~Mahmood, J.~Romero, G.~Pons-Moll, and M.~J. Black, ``Smpl: A
  skinned multi-person linear model,'' \emph{ACM transactions on graphics
  (TOG)}, vol.~34, no.~6, pp. 1--16, 2015.

\bibitem{su2015render}
H.~Su, C.~R. Qi, Y.~Li, and L.~J. Guibas, ``Render for cnn: Viewpoint
  estimation in images using cnns trained with rendered 3d model views,'' in
  \emph{Proceedings of the IEEE International Conference on Computer Vision},
  2015, pp. 2686--2694.

\bibitem{pinheiro2019domain}
P.~O. Pinheiro, N.~Rostamzadeh, and S.~Ahn, ``Domain-adaptive single-view 3d
  reconstruction,'' in \emph{Proceedings of the IEEE/CVF International
  Conference on Computer Vision}, 2019, pp. 7638--7647.

\end{thebibliography}

\begin{IEEEbiography}[{\includegraphics[width=1in,height=1.25in,clip,keepaspectratio]{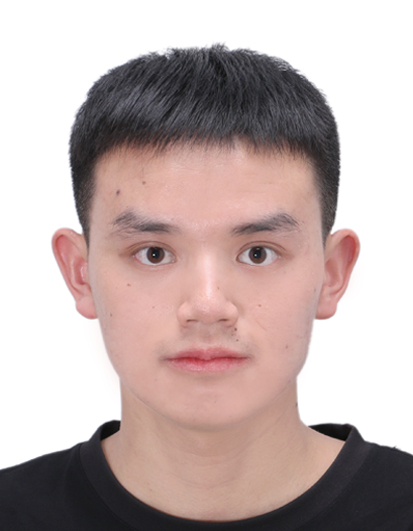}}]
    {Jiapeng Tang} received the B.E. degree in School of Electronic and Information Engineering from South China University of Technology, Guangzhou, China, in 2018, where he is currently pursuing the master’s degree. His current research interests include computer vision and deep learning, especially the 3D reconstruction.
\end{IEEEbiography}

\begin{IEEEbiography}[{\includegraphics[width=1in,height=1.25in,clip,keepaspectratio]{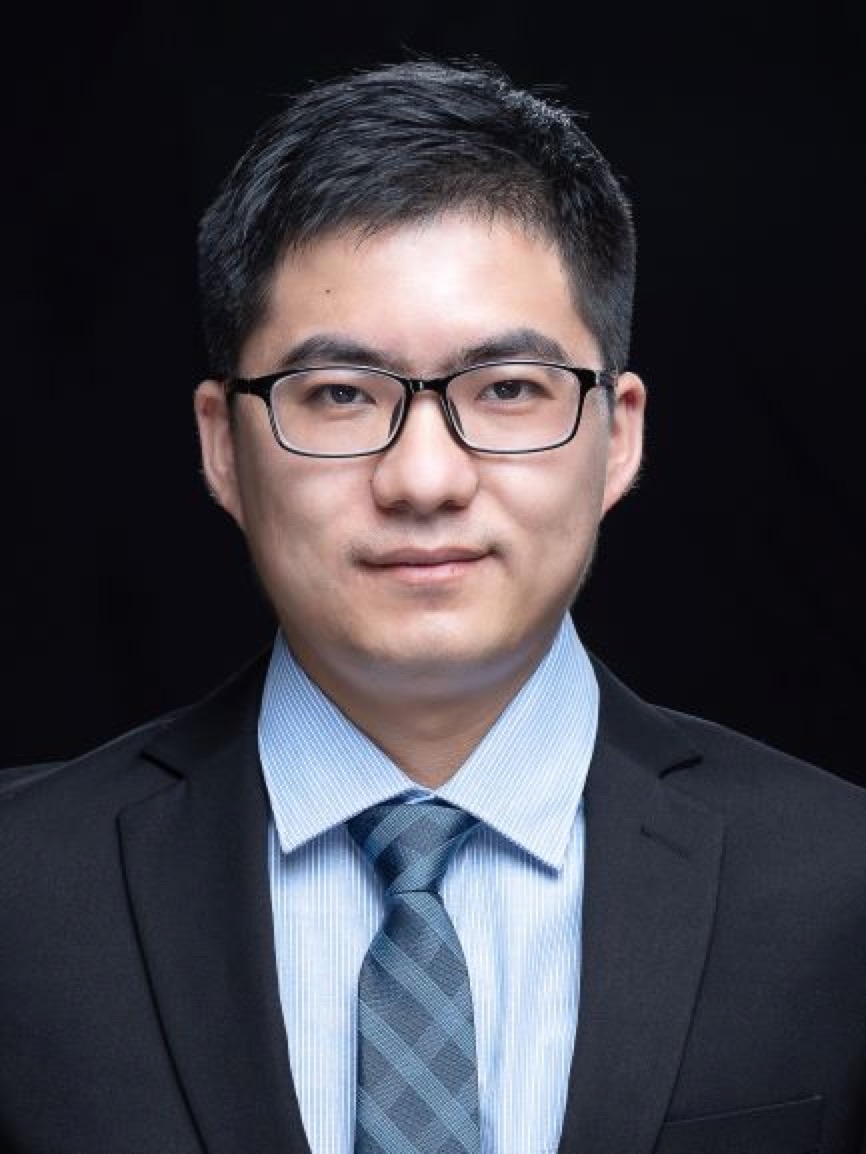}}]
    {Xiaoguang Han} is now an Assistant Professor at The Chinese University of Hong Kong, Shenzhen. He received his Ph.D. degree in computer science from The University of Hong Kong (2013-2017), his M.S. degree in applied mathematics from Zhejiang University (2009-2011) and his B.S. degree in math from Nanjing University of Aeronautics and Astronautics. He also spent 2 years (2011-2013) in City University of Hong Kong as a research associate. His research interests include computer vision, computer graphics, human-computer interaction, medical image analysis, and machine learning.
\end{IEEEbiography}

 
 \begin{IEEEbiography}[{\includegraphics[width=1in,height=1.25in,clip,keepaspectratio]{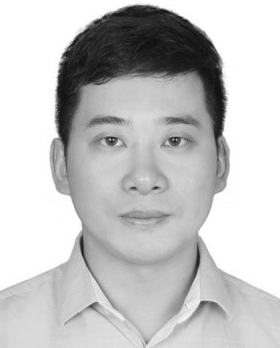}}]
    {Mingkui Tan} received the bachelor's degree in environmental science and engineering and the master's degree in control science and engineering from Hunan University, Changsha, China, in 2006 and 2009, respectively, and the Ph.D. degree in computer science from Nanyang Technological University, Singapore, in 2014. From 2014 to 2016, he was a Senior Research Associate of computer vision with the School of Computer Science, University of Adelaide, Australia. He is currently a Professor with the School of Software Engineering, South China University of Technology. His research interests include machine learning, deep learning, and large-scale optimization.
\end{IEEEbiography}
 
 \begin{IEEEbiography}[{\includegraphics[width=1in,height=1.25in,clip,keepaspectratio]{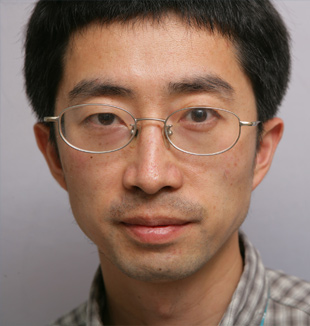}}]
    {Xin Tong} is a principal researcher and leader of the internet graphics group in Microsoft Research Asia. He got his Ph.D. from Tsinghua University in 1999 and then joined Microsoft Research Asia. His research interests are computer graphics and computer vision, including appearance modeling and rendering, computational light transport, texture synthesis, image based rendering, facial performance capturing and modeling, as well as data driven geometric analysis and modeling. Xin Tong got his Bachler and Master degree from Zhejiang University in 1993 and 1996 respectively. He has served as associate editors of IEEE TVCG, CGF, ACM TOG, as well as IEEE CG\&A. Xin is a member of IEEE and ACM.
\end{IEEEbiography}

\begin{IEEEbiography}[{\includegraphics[width=1in,height=1.25in,clip,keepaspectratio]{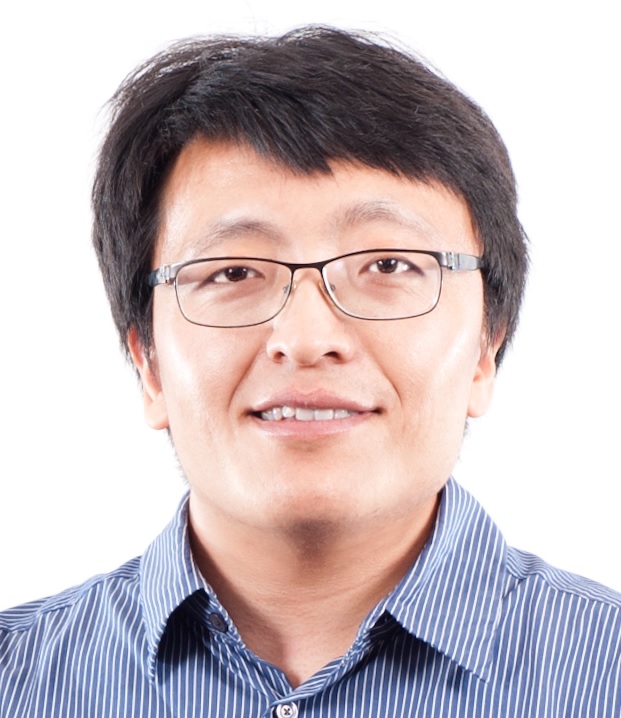}}]
    {Kui Jia} received the B.E. degree from Northwestern Polytechnic University, Xi’an, China, in 2001, the M.E. degree from the National University of Singapore, Singapore, in 2004, and the Ph.D. degree in computer science from the Queen Mary University of London, London, U.K., in 2007. He was with the Shenzhen Institute of Advanced Technology of the Chinese Academy of Sciences, Shenzhen, China, Chinese University of Hong Kong, Hong Kong, the Institute of Advanced Studies, University of Illinois at Urbana-Champaign, Champaign, IL, USA, and the University of Macau, Macau, China. He is currently a Professor with the School of Electronic and Information Engineering, South China University of Technology, Guangzhou, China. His recent research focuses on theoretical deep learning and its applications in vision and robotic problems, including deep learning of 3D data and deep transfer learning. He has been serving as Associate Editors for TIP and TSMC.
 \end{IEEEbiography}


\newpage

\section*{APPENDIX A}
\section*{Discussions on other ShapeNet categories with simple topologies}
    We provide additional qualitative results  on other ShapeNet ~\cite{chang2015shapenet} categories with simple topologies.
    Fig.~\ref{fig:appendix:comp_recon} suggests that our advantages over existing methods become less obvious on those simple objects. Interestingly, from Tables 3 and 4 in the main paper, we can observe that our algorithms are quantitatively superior to existing methods on such simple shapes; this might be due to the designed two-stage pipeline itself that eases the learning of surface meshes. In spite of these observations on simple shapes, we emphasize again that our key motivation of involving skeleton representation is to address the challenges of recovering shapes with complex topologies or thin structures.
    

\section*{APPENDIX B}
\section*{Visualization results of SkeletonNet on the DeepHuman dataset}
\begin{figure}[h]
    	\begin{center}
    		\includegraphics[scale=0.48]{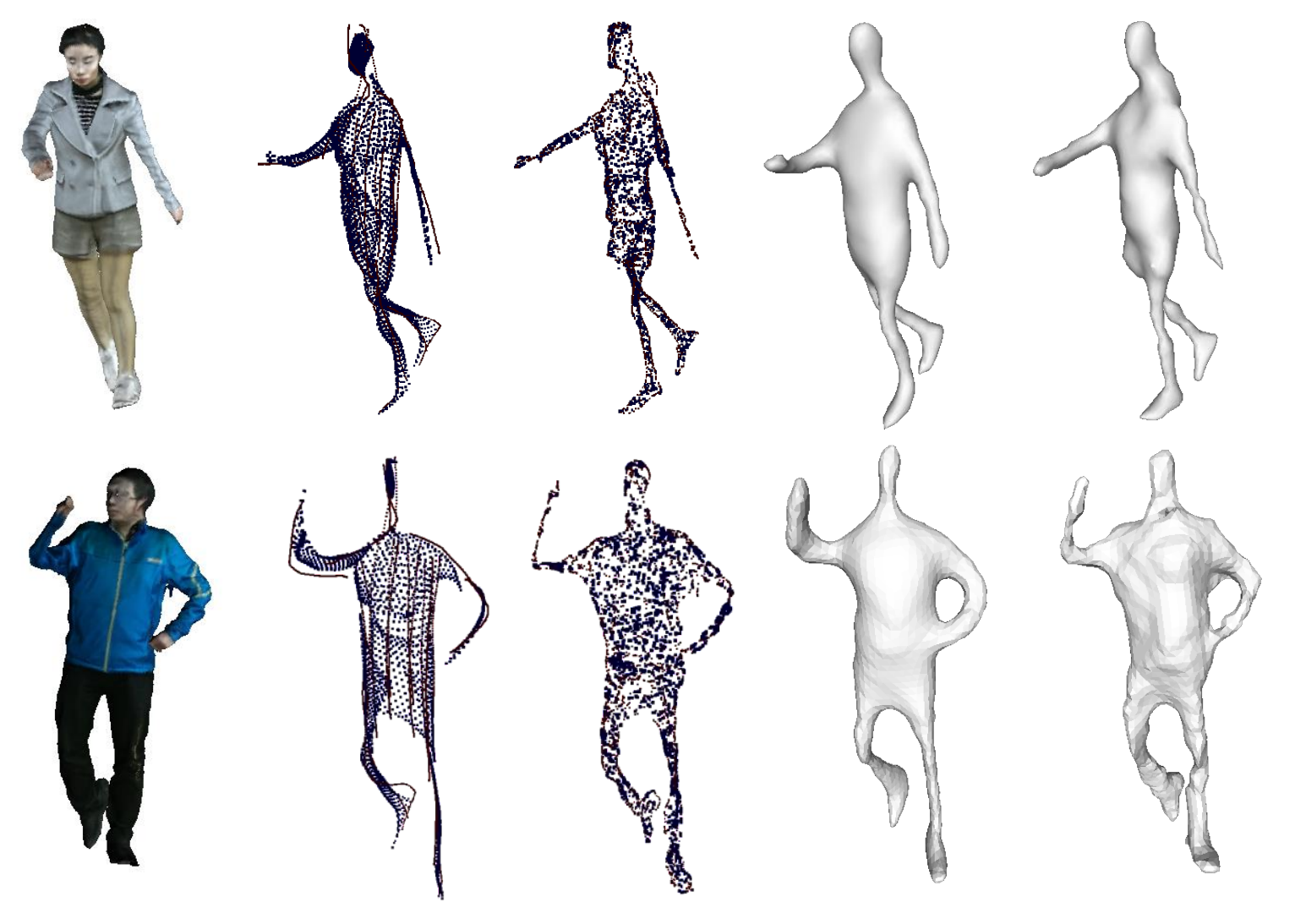}\\
    		\begin{tabular}{p{35pt}p{35pt}p{35pt}p{35pt}p{35pt}}
                \quad \quad  (a)  &
                \quad \quad  (b)  &
                \quad \ \  (c)  &
                \quad \ \  (d) &
                \quad \ \ (e)
            \end{tabular}
    		\caption{Visualization results of our proposed SkeletonNet and its intermediate predictions of skeletal point set on the DeepHuman~\cite{zheng2019deephuman} dataset. (a) Input images; (b) the produced skeletal points; (c) the ground-truth skeletal points; (d) the refined skeletal volumes; (e) the ground-truth skeletal volumes.}
    		\label{fig:skehuman}
    	\end{center}
    	\vspace{-20pt}
\end{figure}

    In Fig.~\ref{fig:skehuman}, we report example results of our proposed SkeletonNet and its intermediate predictions of skeletal point sets on the DeepHuman~\cite{zheng2019deephuman} dataset. The results demonstrate that our SkeletonNet can capture the structures of human shapes with different poses.

\newpage

\end{document}